# IRC-Bench: Recognizing Entities from Contextual Cues in First-Person Reminiscences

**Yehudit Aperstein[1], Eden Moran[1] and Alexander Apartsin [2]**

[1] Intelligent Systems, Afeka Academic College of Engineering, Tel Aviv, Israel;
[2] School of Computer Science, Faculty of Sciences, Holon Institute of Technology, Holon, Israel;

**Abstract**

When people recount personal memories, they often refer to people, places, and events indirectly, relying on contextual cues rather than explicit names. Such implicit references are central to reminiscence narratives: first-person accounts of lived experience used in therapeutic, archival, and social settings. They pose a difficult computational problem because the intended entity must be inferred from dispersed narrative evidence rather than from a local mention. We introduce IRC-Bench, the Implicit Reminiscence Context Benchmark, for evaluating implicit entity recognition in reminiscence transcripts. The benchmark targets non-locality: entity-identifying cues are distributed across multiple, non-contiguous clauses, unlike named entity recognition, entity linking, or coreference resolution. IRC-Bench comprises 25,136 samples constructed from 12,337 Wikidata-linked entities across 1,994 transcripts spanning 11 thematic domains. Each sample pairs an Entity-Grounded Narrative, in which the target entity is explicitly mentioned, with an Entity-Elided Narrative, in which direct mentions are removed. We evaluate 19 configurations across LLM generation, dense retrieval, RAG, and fine-tuning. QLoRA-adapted Llama 3.1 8B performs best in the open-world setting (38.94% exact match; 51.59% Jaccard), while fine-tuned DPR leads closed-world retrieval (35.38% Hit@1; 71.49% Hit@10). We release IRC-Bench with data, code, and evaluation tools.

**Keywords:** implicit entity recognition; reminiscence narratives; coreference resolution; non-locality; large language models; dense passage retrieval; QLoRA; benchmark; entity linking; Wikidata

## 1. Introduction

Reminiscence, the act of recalling and sharing personal memories, plays a central role in human social life. In clinical settings, reminiscence therapy has been shown to reduce depression and improve well-being in older adults [1,2], while in archival contexts, recorded reminiscences preserve cultural and historical knowledge that would otherwise be lost [3]. A defining characteristic of reminiscence narratives is that speakers assume shared context with their audience: they reference people, places, and events through contextual cues rather than explicit naming, trusting the listener to fill in the gaps. This implicit referencing is natural in conversation but creates a fundamental challenge for automated systems that seek to index, search, or analyze these narratives. Table 1 presents a Japanese American reminiscence example used throughout the paper to illustrate the task. It contrasts an entity-grounded narrative (EGN), in which the target entity is still explicitly named, with an entity-elided narrative (EEN), in which explicit names are removed and only contextual cues remain.

**Table 1:** Example of implicit entity recognition. The EGN contains explicit mentions of the target entity, whereas the EEN removes those mentions while preserving contextual cues that allow recovery of the gold entity.

| | |
|---|---|
| ENTITY-GROUNDED NARRATIVE (EGN) | *"The* **attack on Pearl Harbor** *was the event that changed everything for Japanese Americans like me. After December 7, 1941, suspicion and hatred grew, and we were treated as enemy aliens despite being American citizens. It was because of* **Pearl Harbor** *that the government issued Executive Order 9066 and started the forced relocation."* |

**ENTITY-ELIDED NARRATIVE (EEN)**

*"The* **surprise attack on a naval base in Hawaii** *was the event that changed everything for Japanese Americans like me. After December 7, 1941, suspicion and hatred grew, and we were treated as enemy aliens despite being American citizens. It was because of* **that attack** *that the government issued an order and started the forced relocation."*

**Gold entity:** Attack on Pearl Harbor (Q52418) | Type: Event | Cues: December 7, 1941; naval base in Hawaii; Executive Order 9066; forced relocation

A human reader readily identifies the Attack on Pearl Harbor from the constellation of cues: the date, the Hawaiian naval base, the executive order, the internment of Japanese Americans. No single phrase names the entity; instead, recognition depends on integrating cultural, temporal, and historical knowledge distributed across the entire passage. This pattern of *implicit entity reference* is pervasive in reminiscence narratives, where speakers routinely allude to well-known people, places, and events without naming them, relying on shared background knowledge with their listener.

This phenomenon falls between existing NLP tasks without being addressed by any of them. Named Entity Recognition (NER) identifies explicitly mentioned entity spans in text [4,5]. Entity Linking (EL) resolves those spans to knowledge base entries [6,7]. Coreference resolution connects multiple references to the same entity but requires at least one explicit mention as an antecedent [8]. In implicit entity references, the entity is never named anywhere in the text; there is no span to extract, no mention to link, no antecedent to resolve. The task can be viewed as a form of zero-mention coreference: resolving a reference to an entity that has no surface realization in the text, only a distributed constellation of contextual cues.

While implicit entity recognition was first explored in short social-media text [9,10], we extend it to a fundamentally different setting: long-form reminiscence narratives where entity cues are non-local, distributed across multiple clauses. We release IRC-Bench (Implicit Reminiscence Context Benchmark), a large-scale evaluation resource constructed from real reminiscence transcripts. This task addresses practical needs across multiple domains. Archives of personal reminiscences, including oral history collections containing millions of hours of recorded testimony, remain largely inaccessible to structured search because the entities discussed are rarely stated by name [11]. In healthcare, reminiscence therapy is a widely used intervention for older adults with dementia and depression [1,2]; automated systems that support these therapeutic conversations must identify the people and events being discussed even when the speaker does not name them. Social robotics and conversational AI for elderly companionship similarly require understanding implicit references to engage meaningfully with users' personal histories. More broadly, information retrieval over personal narratives requires understanding not just what is said, but what is meant.

Our contributions are as follows:

- **Operationalizing non-locality at scale.** IRC-Bench contributes a concrete operational formulation of non-locality for implicit entity recognition in reminiscence narratives. We define non-locality as a task setting in which the target entity is difficult to infer from any single contiguous passage span, but becomes identifiable through the combination of multiple distributed contextual cues. We operationalize this property through a formal task definition, a sentence-ablation diagnostic, and benchmark-scale evaluation. In the diagnostic experiment, GPT-4o zero-shot accuracy drops from 33.5% with the full text to 12.9% with single-sentence context (n = 200), showing that entity recognition depends on integrating cues distributed across the narrative.

- **IRC-Bench.** We release IRC-Bench as a community evaluation resource: 25,136 implicit entity recognition samples derived from 12,337 unique Wikidata-linked entities sourced from 1,994 reminiscence transcripts across 11 thematic domains, with entity-level train/dev/test splits ensuring zero entity overlap between partitions. Each sample includes both an EGN and an EEN, along with entity metadata (QID, aliases, Wikipedia description). The release follows the NeurIPS Datasets and Benchmarks track convention; contributions (1) and (3) stand independently of the dataset release.
- **Comprehensive evaluation.** We systematically compare 19 experimental configurations spanning open-world LLM inference (zero-shot, few-shot, chain-of-thought, QLoRA fine-tuning), closed-world dense retrieval (off-the-shelf and DPR fine-tuned), and hybrid RAG, revealing that fine-tuning doubles performance in both paradigms, chain-of-thought reasoning degrades performance on this task, and model scale interacts strongly with prompting and adaptation strategy.

## 2. Related Work

### 2.1. Named Entity Recognition

Named Entity Recognition identifies and classifies explicit entity mentions in text. Classical approaches relied on hand-crafted features and conditional random fields [4], while modern systems employ deep learning architectures including bidirectional long short-term memory with conditional random fields (BiLSTM-CRF) [11], transformer-based sequence labeling [12], and large language model prompting [13,14]. Recent benchmarks such as CoNLL-2003 [15] and Multi-CoNER [16] have driven progress across entity types and languages. The W2NER framework [17] unified flat, nested, and discontinuous NER as word-word relation classification, and UniversalNER [18] demonstrated targeted distillation from LLMs for open-domain entity extraction. Despite these advances, all NER formulations assume the target entity appears as an explicit surface form in the input text, an assumption that does not hold for implicit references.

### 2.2 Entity Linking

Entity Linking resolves textual mentions to entries in a knowledge base. Neural approaches include local attention models [6], bi-encoder architectures such as BLINK [19], autoregressive generation via GENRE [20], and efficient zero-shot systems like ReFinED [21]. Botha et al. [22] extended entity linking to over 100 languages. These systems take an identified mention span as input and rank candidate entities; they cannot operate when no mention span exists. Implicit entity recognition requires generating entity candidates from distributed contextual cues rather than resolving a given span.

### 2.3 Reminiscence Analysis and NLP

Reminiscence, the structured recall of autobiographical memories, has been studied extensively in psychology and gerontology. Butler [23] first proposed life review as a therapeutic process, and subsequent work established reminiscence therapy as an evidence-based intervention for depression and cognitive decline in older adults [1,2]. Webster [24] developed the Reminiscence Functions Scale, identifying eight distinct functions of autobiographical memory sharing. Computational approaches to reminiscence have focused primarily on two areas: reminiscence therapy systems and oral history processing. Therapy-oriented systems use conversational agents to elicit and respond to personal memories [25], while oral history processing addresses transcription, topic segmentation, and search [11,26]. However, none of these systems address the fundamental challenge of identifying the entities that speakers reference implicitly. Our work bridges this gap by extending implicit entity recognition, previously studied only in short social-media text [9,10], to the reminiscence domain and providing the first benchmark derived from real reminiscence narratives.

### 2.4 Implicit and Zero-Mention Entities

Limited prior work has addressed entities that are referenced but not named. Hosseini [9] introduced implicit entity recognition in tweets, constructing a dataset of 3,119 tweets with implicit entity mentions. Hosseini and Bagheri [10] developed learning-to-rank methods for this Twitter dataset. Perera et al. [27] explored implicit entity recognition in clinical documents. The coreference resolution community has studied "zero anaphora" and bridging references [28,29], where an entity is referenced indirectly through related concepts.

Our work differs from these efforts in five fundamental ways. First, domain and text structure. Tweets are short (under 280 characters), formulaic, and heavily context-dependent on trending topics; clinical notes follow rigid templates. Reminiscence narratives are extended first-person accounts (typically 50 to 200 words per sample) with rich, diffuse contextual cues spanning dates, locations, personal relationships, sensory details, and historical events. Second, non-locality. In tweets, the implicit entity is typically inferable from a single cue or hashtag context. In reminiscence narratives, building on the bridging-anaphora and zero-mention coreference literature [28, 29] that has long noted indirect reference, we operationalize the non-locality property at scale: a formal definition (no contiguous span suffices to identify the entity, yet the union of non-contiguous cues does), a per-sample sentence-ablation diagnostic, and benchmark-scale evidence that this regime causes systematic failures in standard pipelines. Third, scale and diversity. IRC-Bench contains 25,136 samples spanning 12,337 unique Wikidata-linked entities across 11 thematic domains, compared to 3,119 tweet samples in Hosseini [9] covering primarily entertainment and sports entities. Fourth, entity-level evaluation. We introduce entity-level train/test splitting with zero entity overlap, ensuring that models must generalize to entirely unseen entities rather than memorizing entity-specific patterns. Prior benchmarks used random sample-level splits where the same entity could appear in both training and test data. Fifth, comprehensive method comparison. We systematically evaluate 19 configurations spanning four paradigms (generative LLM, dense retrieval, RAG, fine-tuning), whereas prior work evaluated at most two to three approaches on a single paradigm.

*2.5 Oral History NLP*

Computational analysis of oral histories has received growing attention. Reminiscence-based interventions have been studied in dementia care [1,2], and AI-driven conversational agents have been explored as companions for elderly users [30,31]. Digital storytelling platforms combining AI with augmented reality enable communities to preserve personal narratives [32]. However, these systems primarily facilitate memory recall and do not attempt to recover the implicit entities that speakers reference without naming.

*2.6 Knowledge-Grounded Question Answering*

The closest existing task to implicit entity recognition is knowledge-grounded question answering, where a system must reason over both a text passage and an external knowledge base to produce an answer [33,34]. RAG approaches retrieve relevant knowledge base passages and condition generation on them [35,36]. While implicit entity recognition shares the requirement for external knowledge, it differs in that the "question" is an entire narrative rather than a targeted query, and the answer is always a single entity rather than a free-form text span. Furthermore, implicit entity recognition exhibits the non-locality property: the relevant cues are distributed throughout the passage rather than concentrated near a question token. This structural difference, as we show empirically, causes standard RAG pipelines to underperform direct LLM inference.

*2.7 Recent Developments*

Since the initial study of implicit entity recognition on social media [9, 10], four related threads have advanced the surrounding state of the art without directly addressing the long-form implicit-mention regime. Entity linking with LLMs. Direct prompting frameworks such as ChatEL [44] and LLM-based context augmentation for long-tail entities [45] reformulate entity linking around LLM reasoning, but both presume an explicit surface mention to resolve. Long-context entity tracking. Recent benchmarks stress-test entity and fact recall across thousands of tokens, including BABILong [46] for reasoning-in-a-haystack and RULER [47] for synthetic needle-in-a-haystack variants. The most directly relevant is NoLiMa [48], which shows that long-context retrieval performance collapses when lexical overlap with the query is removed; this is precisely the regime that IRC-Bench targets at the dataset level, with naturally occurring rather than synthetically removed lexical anchors. Coreference and bridging. The CRAC 2023 shared task [49] codifies bridging-adjacent annotation across multilingual corpora but stops short of zero-mention recognition where no antecedent span exists. Oral history NLP. Speech-technology pipelines for archival audio [50] and LLM-based topical and sentiment annotation of oral-history corpora [51] motivate the downstream entity-grounding task we address, none of these existing efforts attempt entity recovery under implicit reference. IRC-Bench occupies the intersection these threads converge on, long-context, lexically de-coupled, entity-grounded recognition over reminiscence narratives.

## 3. Dataset Construction

### *3.1. Overview*

IRC-Bench is constructed through a five-stage automated pipeline that transforms oral history transcripts into implicit entity recognition samples. Each sample consists of a first-person narrative that references a named entity through contextual cues alone, without ever naming it. All pipeline stages use off-the-shelf GPT-4.1-mini via the OpenAI Batch API with deterministic decoding (temperature 0.0 for NER, 0.3 for summarization and rewriting); no fine-tuning is performed for corpus construction. The pipeline leverages GPT-4.1-mini for entity extraction, summary generation, and implicit rewriting, followed by automated validation, producing 25,136 benchmark samples spanning 12,337 unique entities. Verbatim prompts for all stages are released in Appendix A.

### *3.2 Source Collections*

The raw data comprises 1,994 cleaned oral history transcripts drawn from 11 thematic collections. These collections provide broad topical diversity, covering military conflicts, social movements, immigration, public health crises, labor history, and academic life. The first-person narrative style of oral histories naturally provides rich contextual cues (dates, locations, relationships, roles, events) that make implicit entity references solvable for knowledgeable readers, while remaining challenging for automated systems. Table 2 summarizes the source collections and transcript counts used to construct the corpus.

**Table 2.** Source collections for IRC-Bench. Transcript counts reflect cleaned JSON files after the processing pipeline.

| Collection | Transcripts | Sources | Description |
|---|---|---|---|
| Veterans | 517 | Library of Congress VHP, Nevada WWII, Niles Library, Wisconsin Veterans Museum | Military service narratives |
| Immigration | 402 | University of Minnesota, Densho Digital Archive | Immigration and assimilation experiences |
| Regional | 314 | University of Nevada Reno, Kentucky Oral History Commission | Regional and community histories |
| Depression Era | 213 | Federal Writers' Project (Library of Congress) | Great Depression oral histories |
| Japanese American | 156 | Densho Digital Archive | Japanese American internment and post-war |
| Academic | 153 | Columbia University Oral History, Smithsonian Archives of American Art | Academic and university histories |

| Collection | Transcripts | Sources | Description |
|---|---|---|---|
| September 11 | 72 | National Park Service 9/11 Memorial | 9/11 experiences and aftermath |
| Civil Rights | 68 | Civil Rights History Project (Library of Congress) | Civil rights movement narratives |
| COVID-19 | 42 | Various oral history projects | Pandemic experiences |
| Labor | 30 | Labor Archives and Research Center | Labor movement histories |
| Refugee | 27 | Voices of Conscience, UNHCR collections | Refugee experiences |
| **Total** | **1,994** | **11 thematic domains, 25+ institutional archives** | |

*3.3 Pipeline Stages*

The benchmark construction proceeds in five stages, illustrated in Figure 1.

*Stage 1: Transcript Cleaning*. Raw oral history transcripts are cleaned and converted to structured JSON format, preserving the first-person narrative voice while removing interviewer questions and metadata artifacts.

*Stage 2: Named Entity Recognition and Entity Linking.* GPT-4.1-mini extracts entities of seven types (Place, Organization, Person, Event, Work, Military Unit, Other) and, for each, emits the canonical English Wikipedia title (full prompt: Appendix A.6). Each title is resolved to a Wikidata QID using the Wikipedia pageprops.wikibase_item endpoint, with a wbsearchentities fallback on Wikidata for unmatched titles. The resolved QID supplies up to ten aliases (aliases.en) and the Wikidata short description; the first sentence of the matching Wikipedia article is also fetched as description_wiki. Of 12,337 unique entities in the final KB, 84.6% retain a verified Wikipedia URL, 100% retain a Wikidata QID, and 51.2% have at least one alias. This stage produces 31,284 entity mentions across 1,752 transcript files (87.9% coverage).

*Stage 3: Explicit Summary Generation*. For each (transcript, entity) pair, GPT-4.1-mini generates a first-person narrative summary focused on that entity, preserving the contextual cues surrounding the entity's mention (full prompt: Appendix A.7). This produces 25,161 explicit summaries from 1,601 transcript files (80.3% coverage).

*Stage 4: Implicit Rewriting*. For each summary from Stage 3, GPT-4.1-mini produces an entity-elided rewrite under a fixed prompt (Appendix A.8) with five hard rules: (i) remove the entity name and all obvious aliases; (ii) preserve first-person voice; (iii) preserve every contextual cue (dates, places, co-occurring events, roles, people); (iv) add no new information; (v) keep length to 3 to 5 sentences. Entity references are replaced with sentence-local descriptions inferred from the surrounding context (e.g., "Attack on Pearl Harbor" becomes "the attack on the naval base in Hawaii"). Replacement strings are never drawn from the KB. A subsequent deterministic string-match leakage check using each entity's Wikidata alias list filters samples where the name still appears verbatim, removing 25 of 25,161 candidates and yielding the final 25,136 implicit rewrites (80.2% coverage), which form the benchmark's implicit_text field.

*Stage 5: Automated Validation.* The final implicit rewrites undergo automated validation to detect residual entity-name leakage, assess cue sufficiency, and filter narratives for naturalness before inclusion in IRC-Bench.

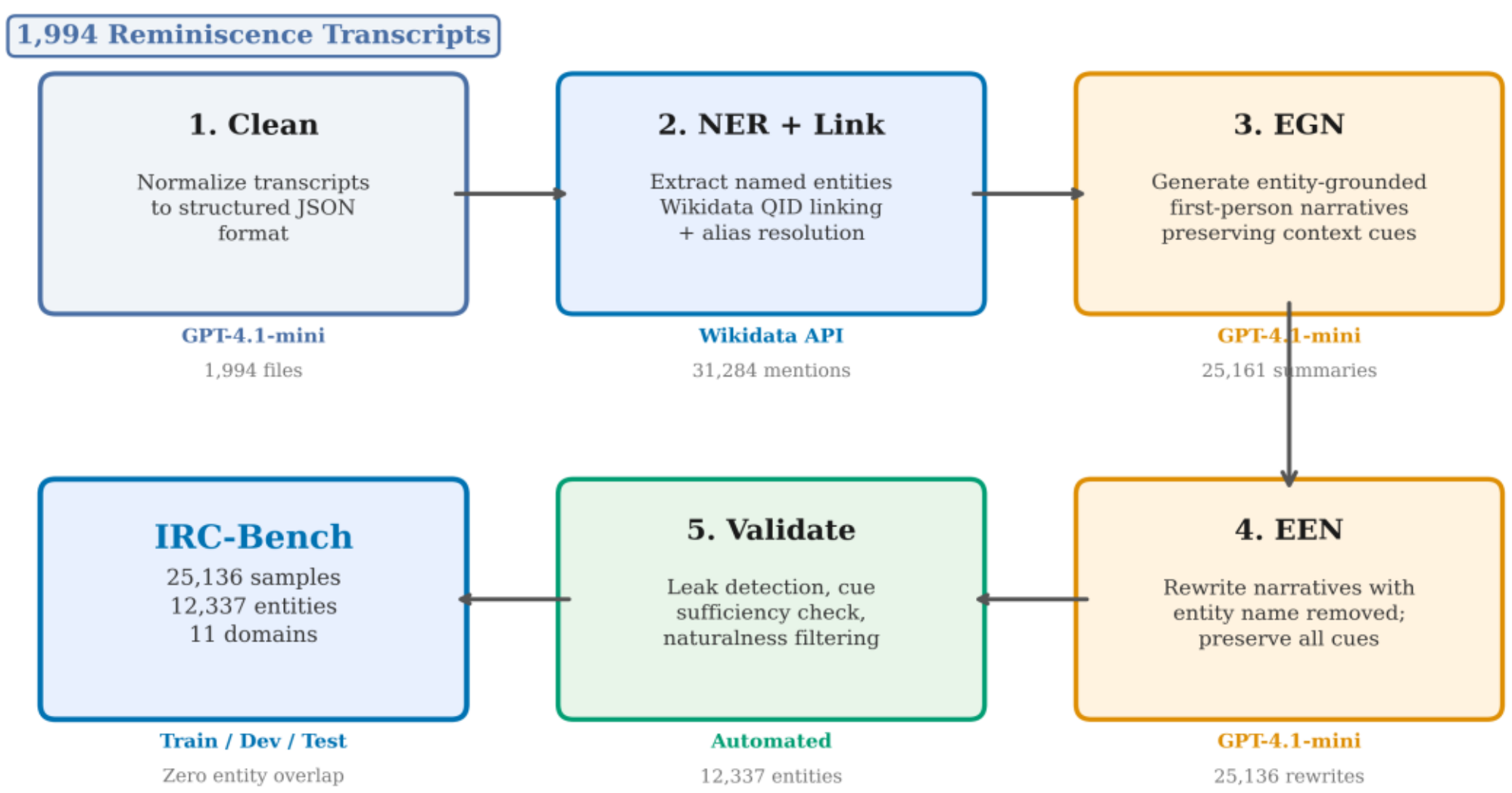

**Figure 1.** IRC-Bench construction pipeline. Raw oral history transcripts undergo cleaning, named entity recognition with Wikidata linking, entity-grounded narrative generation, and entity elision to produce implicit entity recognition evaluation samples. The final validation stage checks for leakage, cue sufficiency, and narrative naturalness.

### *3.4 Entity Knowledge Base*

The entity knowledge base contains 12,337 unique entities with the following metadata coverage: 84.6% have associated Wikipedia pages, 100% have a Wikidata QID, and 51.2% have alternative names sourced from Wikidata. Entity short descriptions used as retrieval targets in the closed-world experiments are sourced from Wikidata (descriptions.en) for 84.6% of entities and from the matching Wikipedia first sentence for an additional 13.7%; for the remaining 1.7% of QID-holding entities with neither, GPT-4.1-mini emits a one-line description from the in-transcript context. These KB descriptions feed the closed-world retrieval baselines only; they never enter an EEN. Entity representations serve three roles: retrieval targets for closed-world experiments, alias sources for evaluation matching, and description inputs for embedding-based approaches.

### *3.5 Entity-Level Train/Dev/Test Splitting*

To ensure rigorous evaluation, the dataset is split at the entity level rather than the sample level. All samples for a given entity appear in exactly one partition, preventing information leakage where a model might learn entity-specific patterns from training examples and exploit them at test time. The split uses a 70/10/20 ratio (seed=42). Table 3 reports the resulting sample and entity counts for each partition.

**Table 3.** IRC-Bench partition statistics. Entity-level splits ensure zero overlap between train, dev, and test entities.

| Partition | Samples | Entities |
|---|---|---|
| Train | 17,971 | 8,635 |
| Dev | 2,532 | 1,234 |
| Test | 4,633 | 2,468 |
| Total | 25,136 | 12,337 |

### *3.6 Entity Type Distribution*

The dataset exhibits a natural long-tail distribution over entity types. Places dominate (47.3%), reflecting oral histories' emphasis on geographic locations. Organizations (21.3%) and Persons (13.7%) follow, while specialized types such as Events, Works, and Military Units are less frequent but still well represented. Table 4 reports the full distribution. Figure 2 visualizes the entity-type distribution alongside the train/dev/test dataset split.

**Table 4.** Distribution of IRC-Bench samples by entity type.

| Entity Type | Samples | % of Total | Unique Entities |
|---|---|---|---|
| Place | 11,893 | 47.3% | 4,821 |
| Organization | 5,366 | 21.3% | 2,894 |
| Person | 3,450 | 13.7% | 2,207 |
| Event | 2,162 | 8.6% | 1,102 |
| Work | 1,195 | 4.8% | 743 |
| Military Unit | 537 | 2.1% | 312 |
| Other | 533 | 2.1% | 258 |
| Total | 25,136 | 100% | 12,337 |

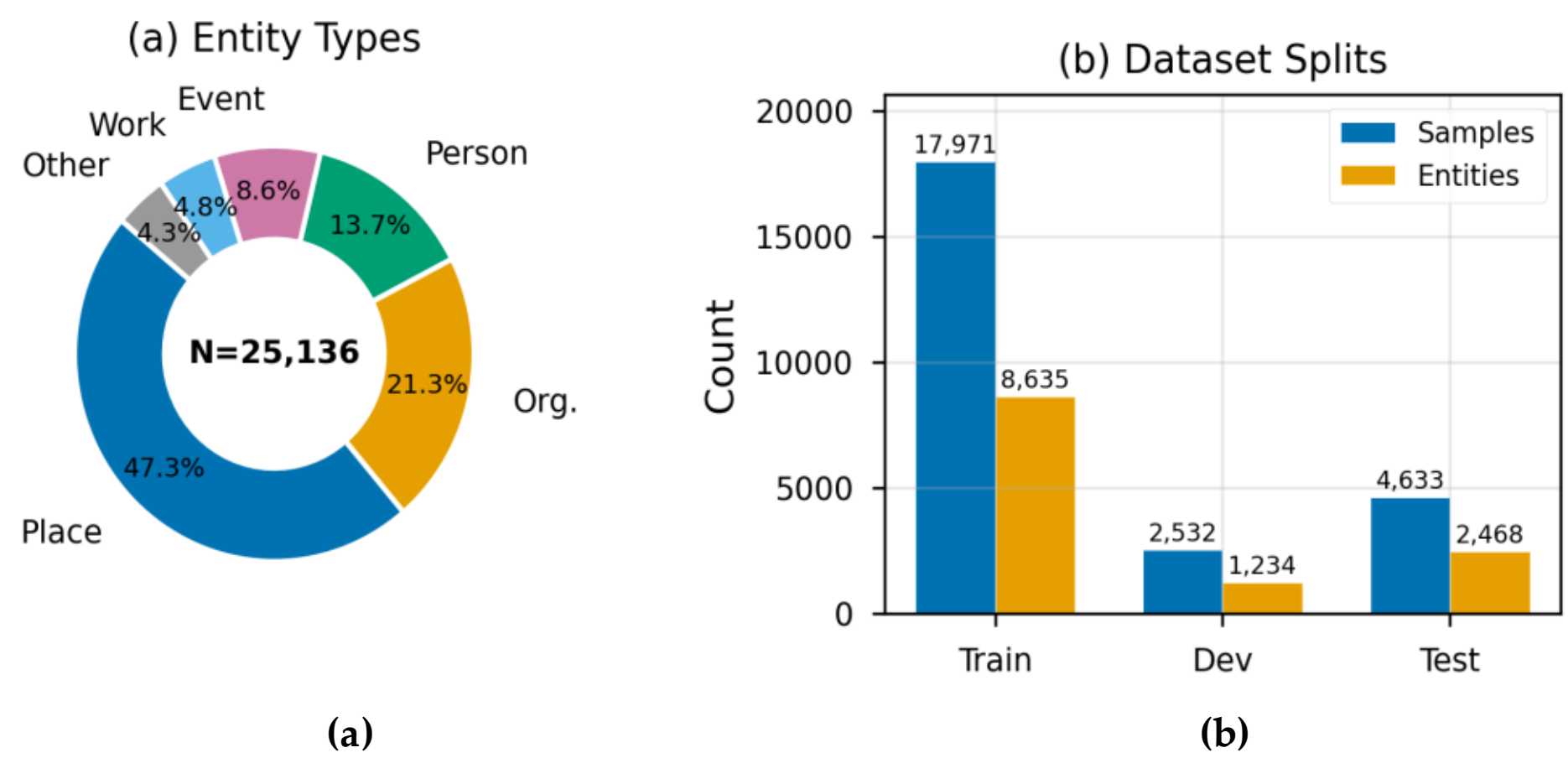

**Figure 2.** IRC-Bench dataset composition. (a) Distribution of samples across entity types. (b) Distribution of samples across train/dev/test splits.

*3.7 Example Samples*

Table 5 presents representative IRC-Bench examples across Person, Event, and Organization entities. The examples demonstrate that the EENs remove explicit entity names while preserving distributed contextual cues that make the target entity recoverable.

**Table 5.** Representative IRC-Bench EGN/EEN examples across entity types. The table illustrates how explicit target-entity mentions in the entity-grounded narrative (EGN) are replaced in the entity-elided narrative (EEN), while contextual cues are preserved to support recovery of the gold entity.

| Entity Type | Example |
|---|---|
| Person | **EGN:** *"***Rosa Parks** *was arrested on December 5, 1955, in Montgomery, Alabama, for refusing to give up her bus seat, an act that sparked the Montgomery bus boycott.* **E. D. Nixon** *called me late that night to inform me of her arrest and to urge action."* <br> **EEN:** *"***A woman** *was arrested on* **December 5, 1955**, *in* **Montgomery, Alabama**, *for refusing to give up her* **bus seat**, *an act that sparked the Montgomery bus boycott.* **A local leader** *called me late that night to inform me of her arrest and to urge action."* <br> **Gold:** Rosa Parks (Q41921) \| **Cues:** December 5 1955, Montgomery Alabama, bus seat refusal, bus boycott |

Event

**EGN**

*"I headed the relief committee during the disastrous* **Berkeley Fire of 1923***, helping to coordinate aid and recovery efforts for the community. This was a challenging time for Berkeley, California, and I took an active role in organizing support to help residents rebuild."*

**EEN**

*"I headed the relief committee during* **the disastrous fire of 1923 in a California city***, helping to coordinate aid and recovery efforts for the community. This was a challenging time for the city, and I took an active role in organizing support to help residents rebuild."*

**Gold:** Berkeley Fire of 1923 (Q4561337)

Organization

**EGN**

*"After leaving the Navy in 1966, I worked in the warehouse at* **Montgomery Ward** *in* **Redwood City***. It was a non-union job and pretty low-key, just me and an older lady doing pricing and warehouse work."*

**EEN**

*"After leaving* **the Navy** *in* **1966***, I worked in the warehouse at* **a national department store** *in* **Redwood City***. It was a* **non-union** *job and pretty low-key, just me and an older lady doing pricing and warehouse work."*

**Gold:** Montgomery Ward (Q3046) | **Cues:** Navy 1966, warehouse, national department store, Redwood City, non-union

### *3.8 Pipeline Validation and Difficulty Calibration*

Before release, Stage-4 outputs are validated on a random sample of 500 test-partition EENs (10.8% of the 4,633-sample test set; 95% binomial CIs of ±4.0 pp at the mean). Each sample is scored by GPT-4o as a structured judge that receives both the entity-grounded narrative (EGN) and the entity-elided narrative (EEN), under a fixed four-dimensional rubric (full prompt: Appendix B.3)

Naturalness on a 1 to 5 Likert scale: 1 = very awkward or robotic; 3 = passable but noticeably stilted; 5 = completely natural first-person speech (intermediate values interpolated).

Leakage (binary, alias-aware): does the entity name or an obvious alias still appear verbatim in the EEN?

Cue sufficiency on 1 to 5: are there enough contextual cues in the EEN for a knowledgeable human to identify the entity?

Recoverability on yes / probably / possible with expertise / unlikely / no: could the judge identify the entity from the EEN alone?

Figure 3 summarizes two complementary analyses. Figure 3a reports the empirical non-locality diagnostic on 200 test samples: GPT-4o zero-shot accuracy drops from 33.5% when given the full EEN to 12.9% when restricted to single-sentence context. This 20.6 percentage-point gap supports the claim that entity recognition in IRC-Bench depends on integrating distributed cues rather than relying on a single local cue. Figure 3b summarizes the benchmark-quality validation on 500 test-partition EENs. Naturalness and cue sufficiency are shown as normalized Likert means, computed as score/5 × 100. Recoverability denotes the percentage of samples judged "yes" or "probably" recoverable, and Best model denotes the highest alias-aware test score rounded to the nearest percentage point.

EEN naturalness averages 4.87 out of 5 (433 fives, 67 fours, none below four), confirming that Stage 4 produces fluent first-person text. In the pre-release validation sample, GPT-4o marked 6.8% of samples (34/500) as possible leakage under the broader judge-based validation rubric. This estimate is reported as a conservative validation signal and is distinct from the deterministic release filter described in Stage 4, which removed 25 of 25,161 full-corpus candidates (0.10%) based on exact string matches against Wikidata aliases. Cue sufficiency averages 3.0 out of 5, indicating moderate overall difficulty with substantial variance.

Recoverability and the cue-imposed ceiling. 42.0% of samples are judged recoverable (5.8% "yes," 36.2% "probably"), 7.2% "possible with expertise," and 50.8% "unlikely" or "no." The 42% rate matches the best system's alias-aware accuracy within 0.6 pp (O10 QLoRA: 41.4%; C5 DPR: 42.8% alias Hit@1), suggesting that top systems approach the practical ceiling imposed by the available cues rather than being bottlenecked by model capacity. The judge prompt and all 500 raw judgments are released for community re-validation.

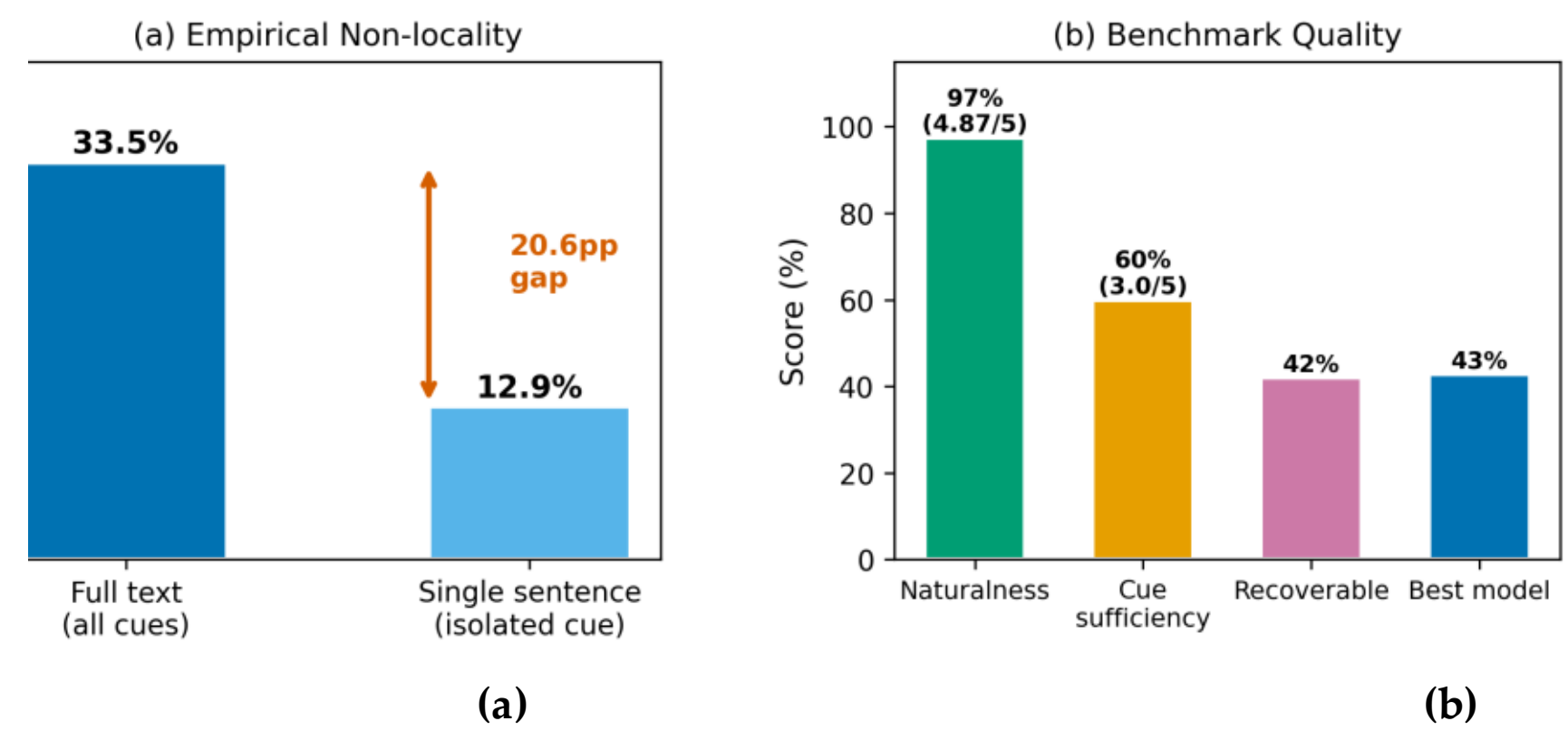

**Figure 3.** Empirical non-locality and benchmark-quality validation. (a) Sentence-ablation diagnostic. (b) Validation summary.

## 4. Methodology

### *4.1. Task Formulation*

Implicit entity recognition, the task of identifying entities that are contextually referenced but never explicitly named, was first studied by Hosseini [9] in the context of tweets. We adopt the same core objective and extend it to long-form reminiscence narratives: given a first-person narrative text $t$ that implicitly references a named entity $e$ without ever mentioning $e$ by name, the task is to identify $e$. The text $t$ contains contextual cues (dates, locations, events, people, roles, descriptions) that jointly constrain the identity of $e$, but the model must synthesize these cues and draw on world knowledge to produce the correct entity name.

We evaluate implicit entity recognition under two formulations:

*Open-world formulation*: The model generates the entity name as free-form text, without access to a candidate set. This tests the model's ability to recall entities from its parametric knowledge. The open-world setting is more realistic, as it does not assume a closed inventory of possible entities.

*Closed-world formulation*: The model ranks all 12,337 entities in the knowledge base by relevance to the query text, selecting the highest-ranked candidate. This tests the model's ability to match implicit descriptions to entity representations via embedding similarity. The closed-world setting provides Hit@K metrics and is analogous to entity linking with a fixed knowledge base.

### *4.2 The Non-Locality Property*

Bridging anaphora [28] and zero-mention coreference [29] have long noted that entities can be referenced indirectly through related concepts; we strengthen this observation by formalizing the case where no contiguous identifying span exists in the entire passage. Let $C(T, e^*) = \{c_1, c_2, \ldots, c_n\}$ denote the set of textual cues in $T$ that collectively identify $e^*$. In standard NER and EL, the entity is localized: there exists a contiguous span $m$ that is sufficient to identify $e^*$. In implicit entity recognition, the entity is *non-local*:

$$\nexists m \subset T \text{ s.t. } m \text{ is contiguous} \wedge m \Rightarrow e^*$$
$$\text{but } C(T, e^*) \Rightarrow e^*, c_i \text{ non-contiguous}$$

That is, no single contiguous substring of $T$ is sufficient to identify $e^*$, but the set of distributed cues collectively determines it. This non-locality has direct implications for method design: approaches that rely on local span matching (NER, EL) or single-vector passage encoding (dense retrieval) are structurally disadvantaged relative to approaches that can integrate information across the full text (LLMs with sufficient context windows).

We empirically validate non-locality by comparing GPT-4o zero-shot accuracy on full implicit texts versus individual sentences in isolation (n=200). Full-text accuracy reaches 33.5%, while single-sentence accuracy drops to 12.9%, a gap of 20.6 percentage points. This confirms that entity recognition requires integrating cues distributed across the entire passage; no single sentence carries sufficient information in the majority of cases.

### *4.3 Open-World Methods*

#### 4.3.1. LLM Generative Approach

We evaluate LLMs in a generative setting where each model receives the implicit text and must produce the entity name. All direct-prompting models use temperature 0.0 (greedy decoding) and a maximum of 100 output tokens. We test zero-shot (ZS) and few-shot (FS, 5 fixed demonstrations) prompting strategies. Few-shot exemplars are selected to cover diverse entity types and are held constant across all test samples. Complete prompt templates appear in Appendix A.

#### 4.3.2 Models

We evaluate three LLM model groups in the open-world setting: GPT-4o [37] and GPT-4.1-mini (via OpenAI Batch API), and Llama 3.1 8B Instruct [38,39] via OpenRouter API. For GPT-4o, GPT-4.1-mini, and Llama 3.1 8B, we additionally evaluate chain-of-thought (CoT) prompting, which instructs the model to reason step-by-step before producing the final answer. CoT experiments use temperature 0.7 and a maximum of 300 output tokens to accommodate the reasoning trace.

#### 4.3.3 QLoRA Fine-tuning (O10)

We fine-tune Llama 3.1 8B Instruct using QLoRA [40,41] for implicit entity recognition. The model is trained to generate the entity name given the implicit text, using the standard causal language modeling objective. The entity-level splitting guarantees zero overlap between training and test entities, so the fine-tuned model cannot memorize entity-specific patterns; it must learn to generalize the implicit-to-entity mapping to entirely unseen entities. Key training parameters include NormalFloat 4-bit (NF4) quantization, LoRA rank 16, alpha 32, learning rate 2e-4, and 2 epochs of training on the full train split (17,971 samples). Full hyperparameters are reported in Appendix B.

### *4.4 Closed-World Methods*

In the closed-world setting, we encode both the implicit query text and all 12,337 entity representations into a shared embedding space, then rank entities by cosine similarity. We explore three entity representation strategies:

*Name* (the entity name alone), *Description* (the entity name concatenated with its LLM-generated description), and *Wiki* (the first sentence from the entity's Wikipedia article). For entities lacking a description or Wikipedia text, we fall back to the next available representation.

4.4.1 BGE-base Baseline (closed-world configurations C1, C2, C3)

We use BAAI/bge-base-en-v1.5 [42] as our baseline embedding model. This 110M-parameter model produces 768-dimensional embeddings and ranks among the top general-purpose bi-encoders on the Massive Text Embedding Benchmark (MTEB). Embeddings are L2-normalized before computing cosine similarity.

4.4.2 DPR Fine-tuning (closed-world configurations C4, C5, C6)

$BB - 1$We fine-tune BGE-base using a DPR approach [36] with Multiple Negatives Ranking Loss (MNRL). Each training pair consists of an implicit text (query) and its gold entity representation (positive passage). MNRL uses in-batch negatives: for a batch of query-positive pairs, each positive for one query serves as a negative for all other queries, providing negatives per sample without explicit hard negative mining. We train for 3 epochs with batch size 48 and learning rate 2e-5. Three separate models are trained, one for each entity representation strategy.

*4.5 RAG Baseline (RAG configuration 1; RAG1)*

We implement a RAG baseline that combines embedding retrieval with LLM reranking. The pipeline operates in two stages. First, BGE-base with entity descriptions (closed-world configuration 2 (C2)) retrieves the top-5 candidate entities for each implicit query. Second, GPT-4.1-mini receives the implicit text along with the 5 candidates (with their descriptions) and selects the most likely entity or suggests a better one. This approach tests whether an LLM can effectively rerank retrieved candidates to improve over pure embedding retrieval.

*4.6 Leakage Prevention*

Information leakage between training and evaluation is controlled at two levels.

*Dataset level.* The benchmark uses an entity-level train/dev/test split (70/10/20, seed=42), verified by set-intersection of entity_list_train.txt, entity_list_dev.txt, and entity_list_test.txt: every pairwise intersection is exactly zero, and every sample for a given entity is assigned to a single partition.

*Model level.* For prompted models (O1 to O8, O11 to O13), the presence of test entities in pretraining is not a leak: entity knowledge is the task itself; only IRC-Bench-specific surface patterns are protected by the split. For QLoRA (O10) and fine-tuned DPR (C4 to C6), the entity-level split guarantees that the gold entity for any test query never appears as a training target or positive. For off-the-shelf retrievers (C1 to C3, RAG1 retrieval), there is no IRC-Bench-specific training. At corpus-construction time, a Stage-4 deterministic string-match leakage check using each entity's Wikidata alias list filtered 25 of 25,161 candidate EENs (0.10%) before release. Separately, the GPT-4o validation audit on 500 sampled test-partition candidates marked 34 cases (6.8%) as possible leakage under a broader judge-based rubric. We therefore report the judge-based leakage rate as a conservative audit signal, while the released corpus filtering itself is defined by the deterministic alias-string procedure. All splits, alias lists, validation judgments, and the leakage-check script are released with the benchmark.

## 5. Evaluation Protocol

*5.1. Matching Hierarchy*

Entity names can be expressed in multiple valid forms (e.g., "United States Marine Corps" vs. "USMC" vs. "Marines"). To account for this variation, we employ a four-tier matching hierarchy, applied in order of decreasing strictness:

*Tier 1 (Exact match)*: The prediction and gold entity are identical after lowercasing and whitespace trimming.

*Tier 2 (Alias match): The prediction matches one of the gold entity's known aliases from Wikidata. For example, predicting "NYC" for gold entity "New York City" is an alias match.*

*Tier 3 (Containment match)*: The prediction is a substring of the gold entity, or vice versa. For example, predicting "Pearl Harbor" for "Attack on Pearl Harbor" qualifies as a containment match.

*Tier 4 (Jaccard match)*: The token-level Jaccard similarity between the prediction and gold entity is at least 0.5. This captures partial overlaps where the prediction includes most of the relevant tokens.

A prediction is considered correct at a given tier if it matches at that tier or any stricter tier. When reporting alias-aware accuracy (the primary metric for open-world experiments), we count any prediction that achieves Tier 1 or Tier 2 as correct.

*5.2 Metrics*

Open-world experiments report exact match (Tier 1), alias match (Tiers 1+2), containment match (Tiers 1+2+3), and Jaccard match (all four tiers). Closed-world experiments report Hit@K (K = 1, 3, 5, 10), Mean Reciprocal Rank (MRR), and alias-aware Hit@1 (where a hit counts if any alias of the gold entity appears in the top-K).

*5.3 Statistical Significance*

To assess whether performance differences between methods are statistically significant, we use McNemar's test (with continuity correction) on the paired per-sample outcomes from each pair of compared systems. Additionally, we compute bootstrap confidence intervals (1,000 resamples, seed=42) at the 95% level.

## 6. Results and Analysis

*6.1 Open-World Performance*

Table 6 presents the open-world results across all experimental configurations. The QLoRA-adapted Llama 3.1 8B (O10) achieves the highest exact match accuracy at 38.94%, substantially outperforming all other open-world methods. Among non-fine-tuned models, GPT-4o with few-shot prompting (O2) is the strongest at 31.62% exact match, rising to 41.10% under the full four-tier Jaccard evaluation. Figure 4 places these open-world results alongside closed-world retrieval performance for comparison.

Model scale is the dominant factor for zero-shot performance: moving from Llama 3.1 8B (13.92%) to GPT-4.1-mini (25.71%) to GPT-4o (27.02%) yields consistent gains. Few-shot prompting consistently improves performance across all model sizes ($p < 0.001$ by McNemar's test). The improvement ranges from +2.95 percentage points for GPT-4.1-mini to +4.60 points for GPT-4o. The few-shot examples appear to calibrate the model's output format and entity granularity, reducing cases where models produce entity types instead of specific entity names.

**Table 6.** Open-world results on the IRC-Bench test set (n=4,633). Exact = exact string match; Alias = alias-aware match; Contain = containment match; Jaccard = Jaccard match (>=0.5). Best result in each column is highlighted.

| ID | Model | Mode | Exact, % | Alias, % | Contain, % | Jaccard, % |
|---|---|---|---|---|---|---|
| O1 | GPT-4o | Zero-shot | 27.02 | 33.30 | 33.30 | 35.05 |
| O2 | GPT-4o | Few-shot | 31.62 | 38.94 | 38.94 | 41.10 |
| O3 | GPT-4.1-mini | Zero-shot | 25.71 | 27.09 | 33.50 | 35.94 |
| O4 | GPT-4.1-mini | Few-shot | 28.66 | 36.89 | 36.89 | 39.48 |
| O5 | Llama 3.1 8B | Zero-shot | 13.92 | 14.81 | 19.47 | 20.18 |
| O6 | Llama 3.1 8B | Few-shot | 17.83 | 18.80 | 24.61 | 25.66 |
| O10 | Llama 3.1 8B (QLoRA) | Fine-tuned | **38.94** | **41.42** | **47.90** | **51.59** |

| ID | Model | Mode | Exact, % | Alias, % | Contain, % | Jaccard, % |
|---|---|---|---|---|---|---|
| O11/b | GPT-4.1-mini CoT | t=0.7 / t=0.0 | 18.93 / 19.44 | 20.27 / 20.76 | 26.48 / 26.87 | 27.69 / 28.10 |
| O12/b | GPT-4o CoT | t=0.7 / t=0.0 | 22.51 / 25.57 | 23.89 / 33.54 | 30.91 / 37.21 | 32.33 / 38.92 |
| O13 | Llama 3.1 8B CoT | t=0.7 | 6.22 | 6.69 | 11.72 | 12.24 |
| RAG1 | BGE + GPT-4.1-mini | RAG | 19.71 | 20.53 | 28.75 | 29.55 |

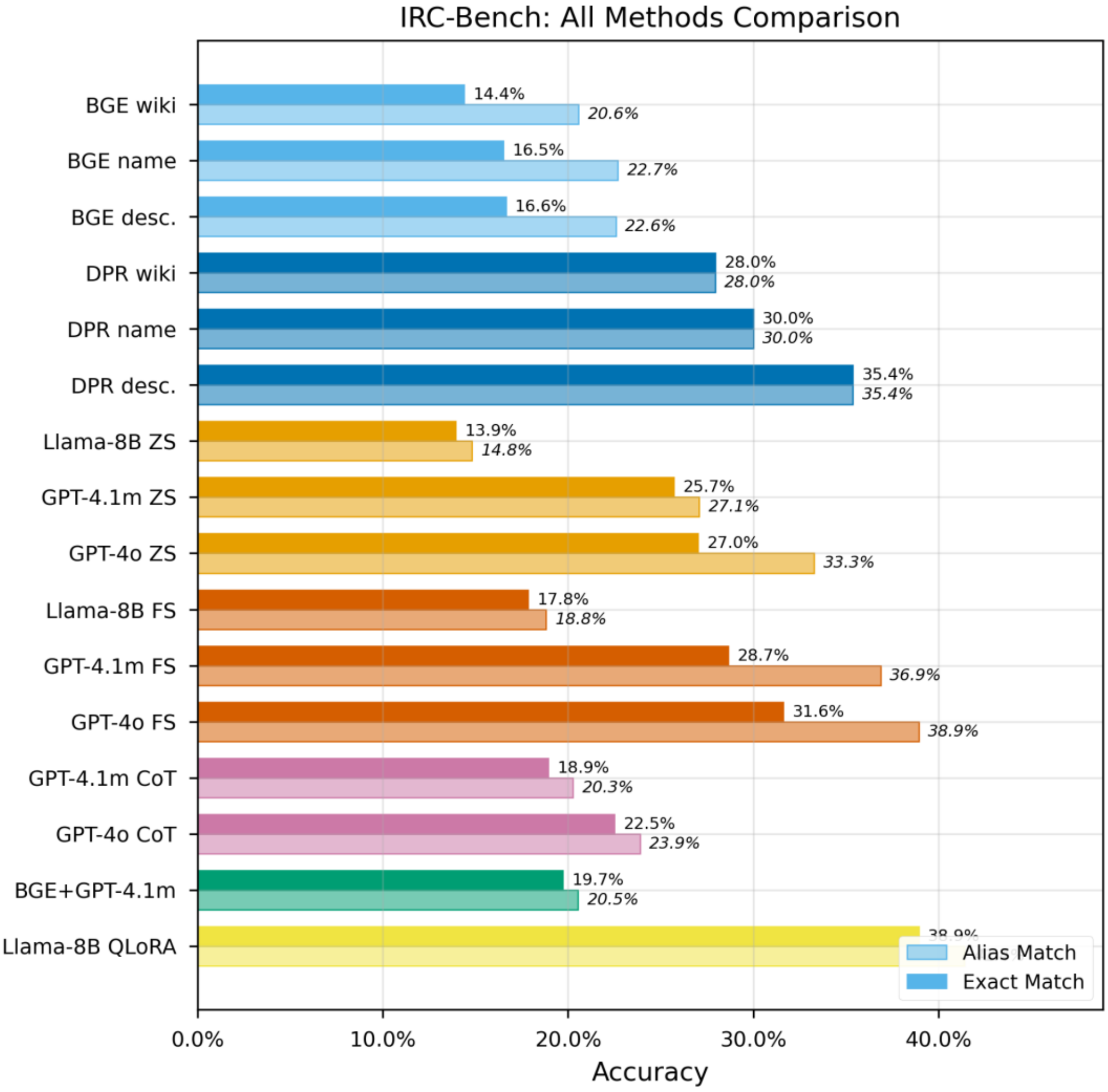

**Figure 4.** Comparison of open-world and closed-world methods on the IRC-Bench test set. For open-world methods, the two bars show exact match and alias-aware accuracy. For closed-world methods, the two bars show Hit@1 and Hit@10.

The most striking open-world result is the effect of QLoRA fine-tuning. Open-world configuration 10 (O10; QLoRA Llama 3.1 8B) achieves 38.94% exact match, nearly tripling the base model's zero-shot performance (13.92%) and exceeding GPT-4o few-shot (31.62%) by 7.32 percentage points. At the Jaccard level, O10 reaches 51.59%, meaning more than half of all test predictions are at least partially correct. This is particularly notable given the entity-level split: O10 has never seen any of the 2,468 test entities during training, demonstrating genuine generalization of the implicit-to-entity mapping.

The *failure of chain-of-thought prompting* is equally striking. CoT reduces GPT-4o accuracy from 33.30% (zero-shot alias) to 23.89%, and GPT-4.1-mini from 25.71% (zero-shot exact) to 18.93%. CoT also degrades Llama 3.1 8B from 13.92% (zero-shot exact) to 6.22%. We analyze the reasons for this failure in Section 7.

The hybrid RAG approach (19.71% exact match) underperforms even GPT-4.1-mini zero-shot (25.71%). When the gold entity does not appear among the top-5 candidates (which occurs in roughly 67% of cases with BGE-base, given C2's Hit@5 of 33.41%), the LLM reranker cannot recover it.

*6.2 Closed-World Performance*

Table 7 shows the closed-world retrieval results. Fine-tuned DPR with description representations (closed-world configuration 5; C5) achieves the best performance: 35.38% Hit@1, 71.49% Hit@10, and 0.4751 MRR. With alias-aware evaluation, C5 reaches 42.80% Hit@1 and 74.47% Hit@10. Figure 5 shows how retrieval performance changes across Hit@K thresholds.

**Table 7.** Closed-world retrieval results on the IRC-Bench test set. The candidate set contains all 12,337 entities. Best result in each column is highlighted. Alias columns report alias-aware metrics.

| ID | Retriever | Entity Repr. | Hit@1, % | Hit@3, % | Hit@5, % | Hit@10, % | MRR | Alias H@1, % |
|---|---|---|---|---|---|---|---|---|
| C1 | BGE (off-the-shelf) | Name | 16.51 | 26.38 | 30.97 | 36.76 | 0.2362 | 22.08 |
| C2 | BGE (off-the-shelf) | Description | 16.64 | 27.78 | 33.41 | 40.60 | 0.2480 | 21.78 |
| C3 | BGE (off-the-shelf) | Wiki | 14.38 | 25.10 | 29.92 | 37.32 | 0.2211 | 19.32 |
| C4 | DPR (fine-tuned) | Name | 30.00 | 46.36 | 53.66 | 63.31 | 0.4131 | 37.10 |
| C5 | DPR (fine-tuned) | Description | **35.38** | **53.51** | **61.82** | **71.49** | **0.4751** | **42.80** |
| C6 | DPR (fine-tuned) | Wiki | 27.95 | 44.98 | 51.82 | 59.55 | 0.3851 | 34.38 |

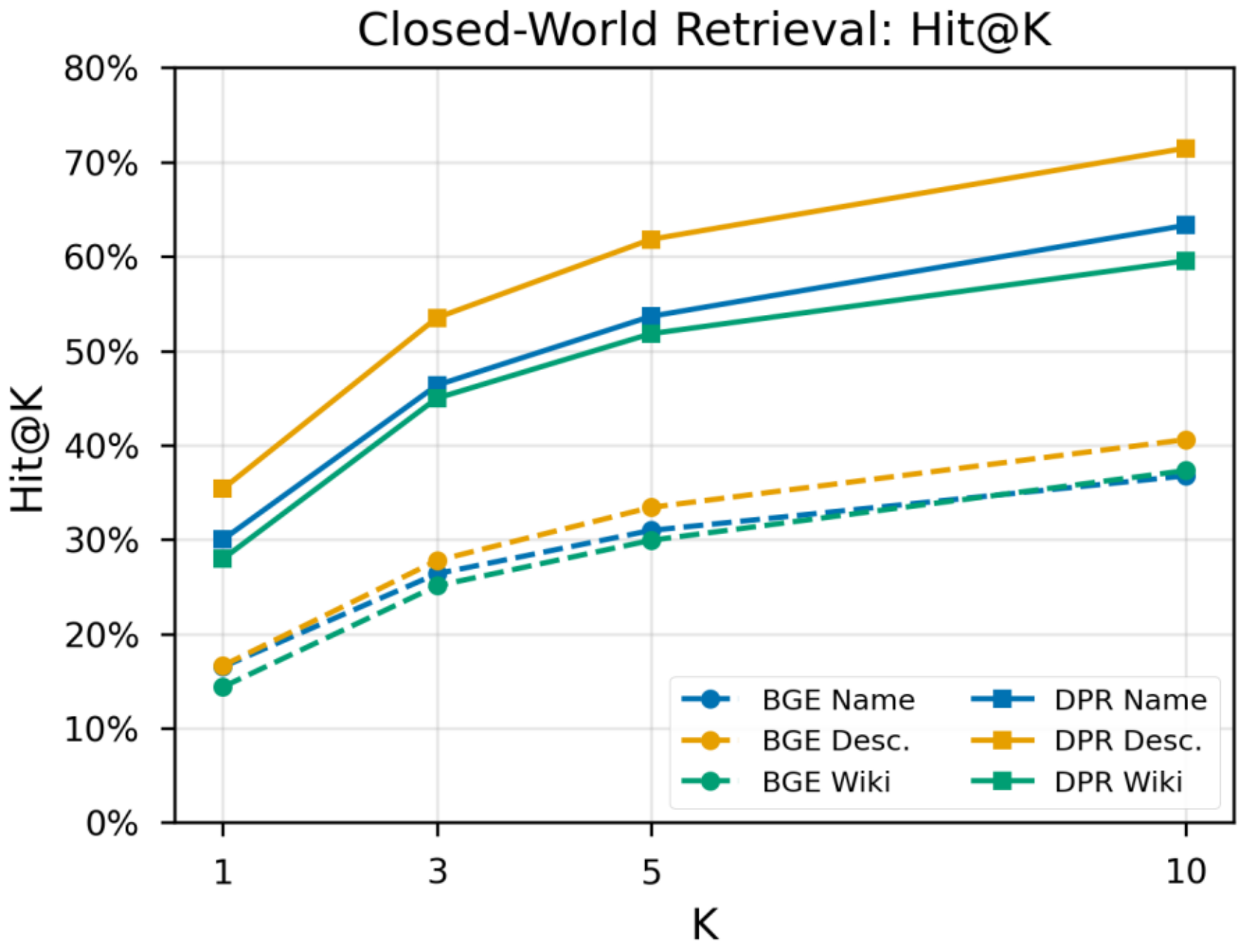

**Figure 5.** Hit@K curves for closed-world retrieval methods. Fine-tuned DPR with description representations (C5) substantially outperforms all baseline configurations across all K values.

The comparison between off-the-shelf BGE and fine-tuned DPR reveals the magnitude of domain adaptation benefits. DPR fine-tuning more than doubles Hit@1 for all entity representation types: Name (16.51% to 30.00%, +13.49 pp), Description (16.64% to 35.38%, +18.74 pp), and Wiki (14.38% to 27.95%, +13.57 pp). The largest absolute gain occurs for descriptions, indicating that fine-tuning is especially effective at learning to align the narrative cue structure with the rich attribute content in entity descriptions. Figure 6 visualizes the effect of DPR fine-tuning across entity representation strategies.

Across both retrieval architectures, entity description representations consistently outperform name-only and Wikipedia representations. Descriptions provide a concise, attribute-rich summary that aligns well with the contextual cues present in elided narratives. Wikipedia lead sentences, despite containing more information, introduce noise from tangential content.

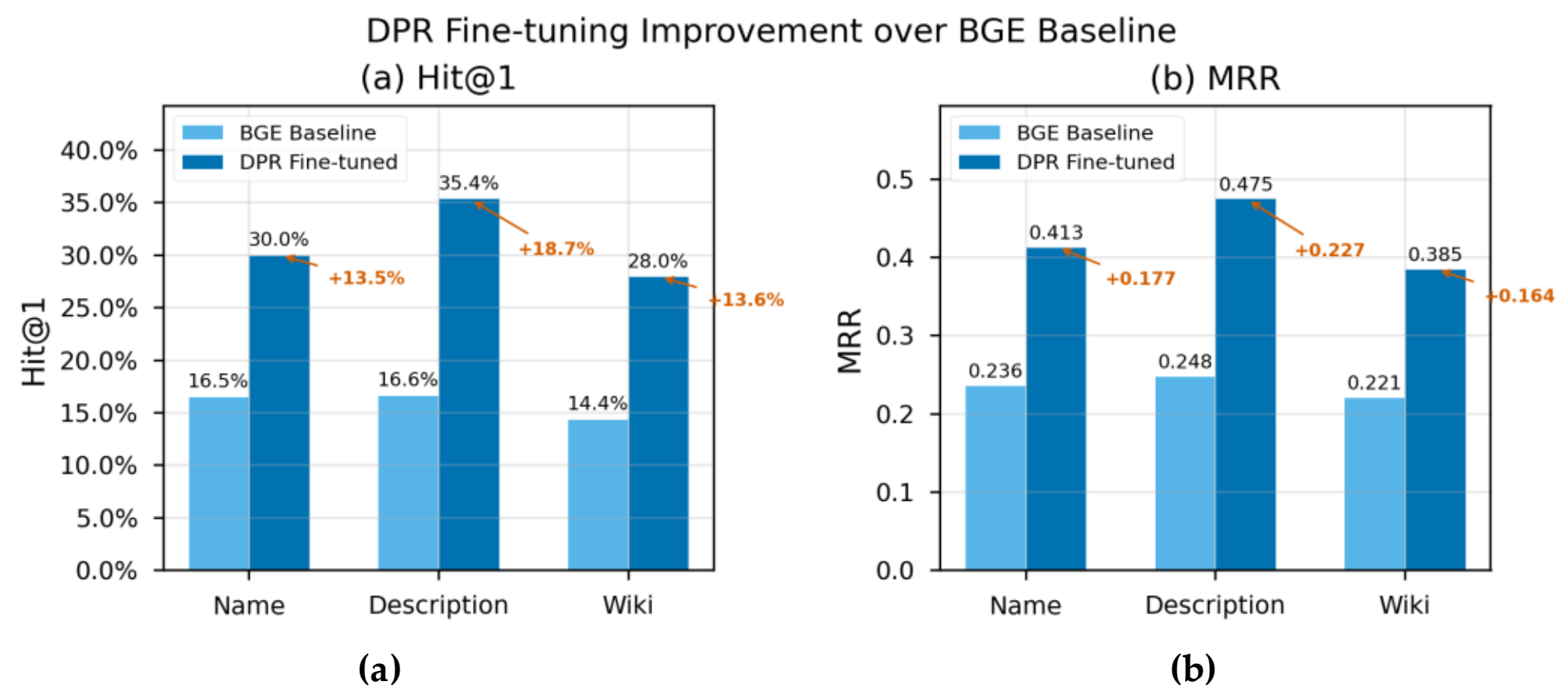

**Figure 6.** Effect of DPR fine-tuning on retrieval performance. Fine-tuning more than doubles Hit@1 across all entity representation strategies, with the largest absolute gain for descriptions (+18.74 pp).

### 6.3 Cross-Paradigm Comparison

Table 8 ranks the top-performing systems across both paradigms under an alias-level com-parison metric.

**Table 8.** Cross-paradigm ranking by alias-level score. Open-world methods use the Alias, % score (Tiers 1+2); closed-world methods use alias-aware Hit@1.

| Rank | System | Paradigm | Alias-level score, % |
|---|---|---|---|
| 1 | C5 (DPR + Description) | Closed | 42.80 |
| 2 | O10 (QLoRA Llama 8B) | Open | 41.42 |
| 3 | O2 (GPT-4o FS) | Open | 38.94 |
| 4 | C4 (DPR + Name) | Closed | 37.10 |
| 5 | O4 (GPT-4.1-mini FS) | Open | 36.89 |
| 6 | C6 (DPR + Wiki) | Closed | 34.38 |
| 7 | O1 (GPT-4o ZS) | Open | 33.30 |
| 8 | O3 (GPT-4.1-mini ZS) | Open | 27.09 |

Under this stricter alias-level comparison, the fine-tuned DPR retriever with descrip-tion representations (C5) ranks first at 42.80% alias-aware Hit@1, followed closely by the fine-tuned QLoRA model (O10) at 41.42% open-world alias accuracy. GPT-4o few-shot (O2) ranks third at 38.94%, and DPR with name representations (C4) ranks fourth at 37.10%. This comparison shows that the 110M-parameter fine-tuned retriever is competitive with, and slightly ahead of, the best open-world generative configuration under an alias-level criterion.

### 6.4 Per-Entity-Type Analysis

Performance varies substantially by entity type. Table 9 reports alias-aware performance by entity type for selected methods. Figure 7 visualizes these per-type differences as a heatmap.

**Table 9.** Hit@1 (%) by entity type for selected methods. Open-world values use alias-aware accuracy; closed-world values use alias-aware Hit@1. n denotes the number of test samples of each type.

| Entity Type | n | O1 (GPT-4o ZS) | O2 (GPT-4o FS) | O5 (Llama 8B ZS) | C1 (BGE Name) | C2 (BGE Desc) |
|---|---|---|---|---|---|---|
| Place | 2,076 | 38.15 | **43.88** | 18.16 | 14.88 | 15.99 |
| Organization | 1,152 | 38.28 | **45.31** | 27.34 | 29.17 | 27.17 |
| Person | 698 | 23.82 | 24.07 | 14.90 | 18.34 | 18.62 |
| Event | 273 | 34.43 | **50.18** | 27.11 | 48.35 | 47.25 |
| Work | 215 | 32.09 | 39.53 | 14.42 | 39.07 | 36.74 |
| Military Unit | 121 | 26.45 | 37.19 | 10.74 | 23.97 | 31.40 |
| Other | 98 | 30.61 | 36.73 | 21.43 | 33.67 | 25.51 |

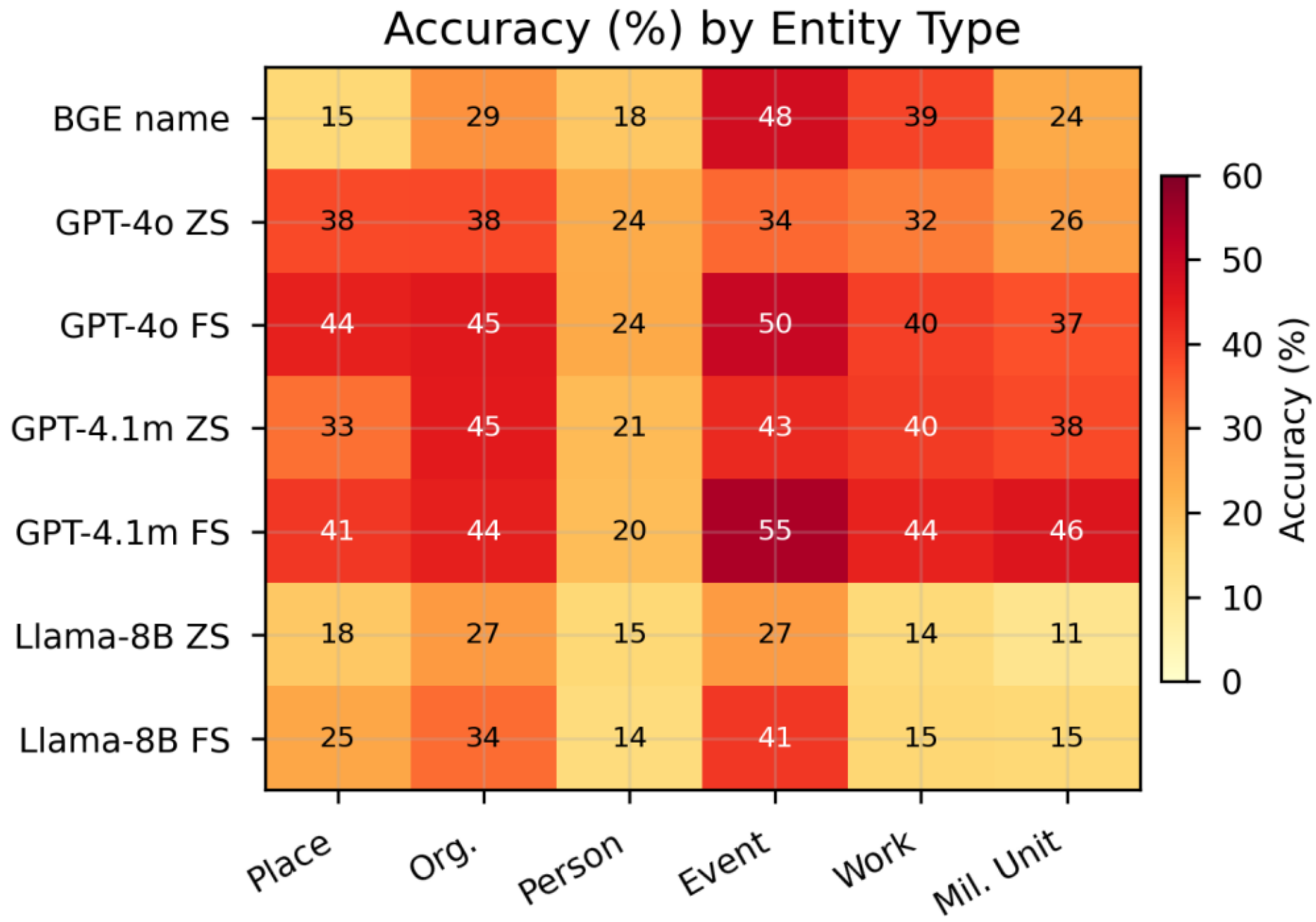

**Figure 7.** Heatmap of performance (alias-aware Hit@1) by entity type and method. Person entities are consistently the hardest across all methods; Events are notably strong for both open-world and closed-world approaches.

**Persons are the hardest type for open-world methods.** GPT-4o FS achieves only 24.07% on Person entities, compared to 45.31% on Organizations and 43.88% on Places. Person entities often have less distinctive contextual cues and are more likely to be obscure individuals not well represented in model training data.

**Events are notably strong for closed-world methods.** BGE achieves 48.35% Hit@1 on Events, higher than any other type, suggesting that event descriptions provide distinctive semantic signatures that align well with implicit event narratives.

**Few-shot examples disproportionately help Events.** GPT-4o jumps from 34.43% (ZS) to 50.18% (FS) on Events (+15.75 pp), the largest per-type improvement, likely because the few-shot examples include two Event instances (Attack on Pearl Harbor).

*6.5 Error Analysis*

We performed automated error classification on 200 randomly sampled incorrect predictions from each of O1 through O6, using GPT-4.1-mini to categorize errors. Table 10 reports the distribution. Representative correct and incorrect prediction examples are provided in Appendix C.

**Table 10.** Error type distribution (%) over 200 randomly sampled incorrect predictions per model. Categories are mutually exclusive.

| Error Type | O1 | O2 | O3 | O4 | O5 | O6 |
|---|---|---|---|---|---|---|
| Same-type, unrelated | 43.0 | 42.0 | 43.5 | 45.0 | 52.0 | 46.0 |
| Wrong type | 28.5 | 27.5 | 29.5 | 22.5 | 31.0 | 35.0 |
| Same-type, related | 24.5 | 25.5 | 22.5 | 24.0 | 13.5 | 17.0 |

| Error Type | O1 | O2 | O3 | O4 | O5 | O6 |
|---|---|---|---|---|---|---|
| Partial match | 3.5 | 4.0 | 3.0 | 6.0 | 2.5 | 1.5 |
| Empty/hallucination | 0.5 | 1.0 | 1.5 | 2.5 | 1.0 | 0.0 |

The dominant error mode across all models is *same-type, unrelated* (42% to 52%), where the model predicts an entity of the correct type but one that is semantically unrelated to the gold entity (e.g., predicting "Jack Johnson" when the gold is "Lou Ambers," both boxers). The second most common error is *wrong type* (22.5% to 35.0%), where the model predicts an entity of an entirely different category. *Same-type, related* errors (13.5% to 25.5%) represent near-misses where the prediction is semantically close to the gold (e.g., predicting "Okinawa" for "Iwo Jima"). Hallucinations and empty responses are rare (<2.5%), indicating that models reliably produce plausible entity names even when incorrect.

Llama 3.1 8B (O5, O6) shows a higher proportion of same-type, unrelated errors (52.0% and 46.0%) and a lower proportion of same-type, related errors (13.5% and 17.0%) compared to GPT models (O1, O2: 24.5% and 25.5%). This suggests that smaller models have weaker ability to narrow down candidates within a type using fine-grained contextual cues.

### *6.6 Key Findings Summary*

Table 11 consolidates the main empirical findings and links each conclusion to the corresponding quantitative comparison.

**Table 11.** Main empirical findings and supporting comparisons.

| # | Claim | Experimental Results |
|---|---|---|
| 1 | Fine-tuning is the most impactful intervention | QLoRA fine-tuning of Llama 3.1 8B raises exact match from 13.92% (O5, zero-shot) to 38.94% (O10), a 2.80x improvement. DPR fine-tuning of BGE raises Hit@1 from 16.64% (C2) to 35.38% (C5), a 2.13x improvement. Both gains are achieved despite zero entity overlap between training and test sets. |
| 2 | QLoRA fine-tuning yields the overall best performance | O10 achieves 38.94% exact match (51.59% Jaccard), surpassing GPT-4o few-shot (31.62% exact, 41.10% Jaccard) by 7.32 pp on exact match and 10.49 pp on Jaccard. This result uses only 6.5M trainable parameters on top of an 8B-parameter base. |
| 3 | Chain-of-thought does not improve performance | CoT reduces GPT-4.1-mini from 25.71% (zero-shot exact) to 18.93% and Llama 3.1 8B from 13.92% to 6.22%. For GPT-4o, lowering temperature recovers parity with zero-shot but does not exceed it. |
| 4 | Few-shot prompting consistently helps | Adding five demonstrations improves GPT-4o from 27.02% to 31.62% (+4.60 pp), GPT-4.1-mini from 25.71% to 28.66% (+2.95 pp), and Llama 3.1 8B from 13.92% to 17.83% (+3.91 pp). |
| 5 | Entity descriptions are the best retrieval representation | C5 (DPR + description) outperforms C4 (DPR + name) by 5.38 pp on Hit@1 (35.38% vs. 30.00%) and C6 (DPR + wiki) by 7.43 pp (35.38% vs. 27.95%). The same pattern appears with off-the-shelf BGE. |
| 6 | RAG underperforms direct LLM inference | RAG1 reaches 19.71% exact match, which is 5.99 pp below GPT-4.1-mini zero-shot (25.71%) and 8.95 pp |
| | | below GPT-4.1-mini few-shot (28.66%). The retrieval bottleneck limits the reranking stage. |
| 7 | Model scale matters in the zero-shot regime | GPT-4o zero-shot reaches 27.02% exact match, outperforming Llama 3.1 8B zero-shot at 13.92% by 13.10 pp (McNemar chi-squared = 432.28, $p < 0.001$). |
| 8 | Retriever Hit@10 reveals strong latent signal | C5 places the gold entity in the top 10 for 71.49% of queries, indicating that DPR shortlists contain useful candidates and can support stronger future reranking approaches. |

*6.7 Statistical Significance*

All key comparisons are statistically significant at $p < 0.001$ (McNemar's test with continuity correction). Table 12 reports the detailed results.

**Table 12.** Statistical significance tests (McNemar's test with continuity correction). All p-values < 0.001. A-only and B-only report the number of discordant pairs.

| Comparison | Acc A (%) | Acc B (%) | McNemar $\chi^2$ | A-only | B-only |
|---|---|---|---|---|---|
| O1 vs O2 (ZS vs FS, GPT-4o) | 35.06 | 41.11 | 149.69 | 120 | 400 |
| O3 vs O4 (ZS vs FS, mini) | 35.16 | 38.72 | 36.18 | 150 | 275 |
| O1 vs O5 (GPT-4o vs Llama 8B) | 35.06 | 20.19 | 432.28 | 892 | 203 |
| O1 vs C2 (Open vs Closed) | 35.06 | 22.58 | 181.93 | 1,204 | 626 |

The discordant pair counts are informative: for the GPT-4o vs. Llama 8B comparison, 892 samples are solved only by GPT-4o while only 203 are solved only by Llama 8B, demonstrating a strong directional advantage. For the ZS vs. FS comparison on GPT-4o (O1 vs. O2), 400 samples are gained while only 120 are lost, confirming that few-shot examples provide a net benefit with limited trade-offs. The 95% bootstrap confidence intervals confirm non-overlapping ranges for all reported comparisons.

## 7. Discussion

*7.1 Implications for system design*

Three actionable consequences follow from our results. *(i) RAG bottleneck.* The RAG pipeline underperforms direct generation (19.71% vs 28.66% for GPT-4.1-mini FS) because dense retrievers encode the EEN as a single vector, losing fine-grained cue information; retrieved candidates are topically related but often incorrect, and when presented as context they can override the model's own correct intuition. With BGE-base, the gold entity appears in the top-5 only 33.41% of the time, severely limiting the reranker. Systems indexing oral-history archives over Wikidata-aligned KBs should expect this ~9 pp penalty unless they fine-tune the retriever. *(ii) Disable CoT.* The failure of chain-of-thought is the most counterintuitive finding. CoT improves mathematical reasoning and multi-hop QA by decomposing problems [32], yet it degrades implicit entity recognition for every model tested because identifying an implicit entity requires simultaneously attending to a constellation of distributed cues; when forced to reason step by step, models fixate on individual cues in isolation, arriving at locally plausible but globally incorrect entities. Temperature control experiments (O11b, O12b) confirm this is structural for smaller models (GPT-4.1-mini: +0.5pp at t=0.0, still 6.3pp below ZS); for GPT-4o, controlling temperature recovers parity (33.5% vs 33.3%) without exceeding it. CoT therefore neither helps nor hurts large models once temperature is controlled, but genuinely harms smaller models. *(iii) QLoRA provides a cost-effective adaptation path.* A short QLoRA pass on a domain corpus closes a 25 pp gap against a frontier closed model at under 20 USD of GPU cost, materially changing the cost curve for building entity-grounding components for archives.

*7.2 Generalization and boundary conditions*

The non-locality property is not inherently tied to English or to oral-history data, since it follows from how speakers use distributed contextual cues to refer to entities. Our empirical claims are bounded by language (English), narrator perspective (first-person reminiscence), entity universe (Wikidata-linkable), and domain (US-centric oral histories). We expect the pipeline to transfer to non-English memoirs and to clinical case histories, but cue density on celebrity entities (over-represented in pre-training) and on hyper-local entities (under-represented) sets the realistic accuracy band; the per-entity-type spread (Persons 24%, Events 50%) is a within-domain reflection of this, driven by cue specificity and KB-neighborhood density. The success of QLoRA (O10: 38.94% exact, up from 13.92% base) despite zero entity overlap indicates that fine-tuning teaches three transferable skills, not memorization: the task format (extracting a single canonical name), cue integration patterns (which temporal, spatial, and relational combinations are diagnostic), and entity type priors (calibrating expectations to reduce wrong-type errors). The model learns "how to solve implicit entity puzzles" rather than memorizing specific answers. Figure 8 visualizes the relationship between approximate model scale, prompting strategy, and open-world exact-match accuracy. The figure shows that larger closed models tend to improve over smaller untuned models, but performance is not determined by scale alone: QLoRA adaptation enables the 8B Llama model to outperform substantially larger closed models. Comparing the best open-world (O10: 38.94%) and closed-world (C5: 35.38% Hit@1, 42.80% alias) results, both paradigms reach comparable alias-level performance, while C5's 71.49% Hit@10 suggests that combining fine-tuned retrieval shortlists with fine-tuned LLM reranking is a promising future direction.

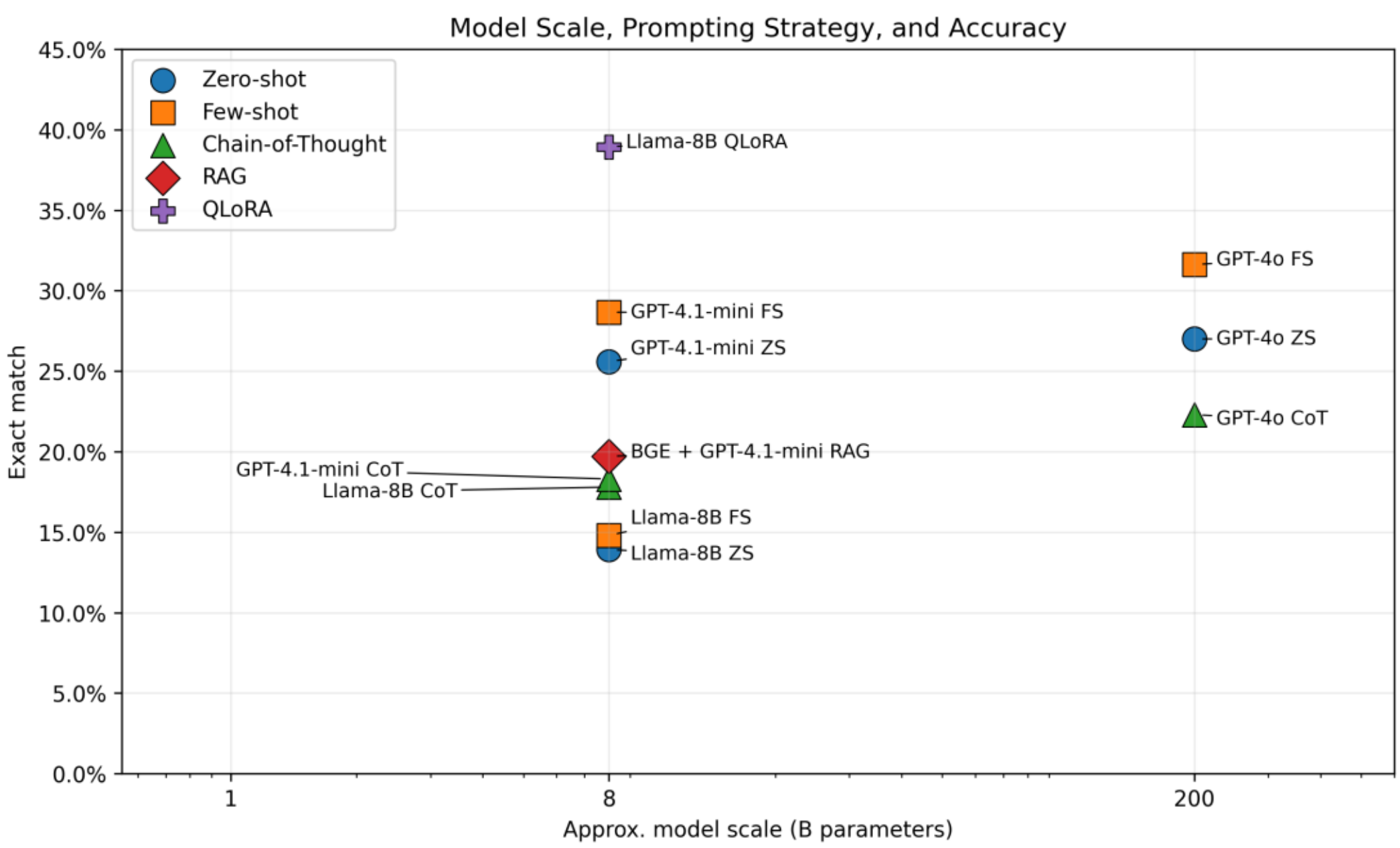

**Figure 8.** Model scale, prompting strategy, and open-world exact-match accuracy. The figure compares zero-shot, few-shot, chain-of-thought, RAG, and QLoRA configurations across approximate model-scale groups, showing that adaptation strategy can outweigh model size.

*7.3 Application areas*

IRC-Bench supports four downstream application areas. *(i) Digital humanities:* automatic entity-grounded indexing of the $10^5$-order oral history corpora curated by national archives. *(ii) Reminiscence therapy systems:* grounding implicit references in dialogue to retrieve relevant photos or documents [25]. *(iii) Privacy-preserving de-identification:* the Stage-4 EENs are a controlled testbed for whether an "anonymized" narrative remains re-identifiable from contextual cues alone, with policy implications for archive release. *(iv) LLM knowledge evaluation:* IRC-Bench measures knowledge of marginalized, regional, and historical entities under-represented in QA benchmarks, exposing gaps that MMLU-style evaluations do not.

*7.4 What IRC-Bench measures that prior tasks do not*

NER measures span localization; entity linking measures name-to-KB resolution; coreference resolution measures intra-document anaphora given at least one explicit mention. IRC-Bench measures *knowledge-grounded cue integration in the absence of any antecedent*, complementing rather than substituting for any of the three. The non-locality ablation (33.5% full-text vs. 12.9% single-sentence, §4.2) is the empirical operationalization of this distinction; we recommend it as a per-task diagnostic for new datasets that claim implicit-reference status.

*7.5 Threats to validity*

*LLM-family overlap.* The pipeline uses GPT-4.1-mini (Stage 4 rewriter), GPT-4o (judge), and a mix of OpenAI and Llama models as evaluation targets. The strongest evidence against systematic family bias is that the best system on IRC-Bench is O10, a QLoRA-adapted Llama 3.1 8B at 38.94% exact match, exceeding GPT-4o few-shot (31.62%) by 7.32 pp; if our pipeline favored OpenAI models, GPT-4o would top the leaderboard, which it does not. The judge (GPT-4o) differs in training mix and class from the generator (GPT-4.1-mini), and the leakage check is an objective alias-string match independent of judge subjectivity. The DPR retrieval line (BGE-base, a Hugging Face encoder) shows the same per-type ordering as the LLM line (Events easiest, Persons hardest), an independent cross-check that the difficulty signal is not GPT-specific.

*AI-only validation.* We acknowledge the absence of human ratings on the calibration set as a limitation, and release the GPT-4o judge prompt and 500 raw judgments for community re-validation. The strongest argument against LLM-judge bias affecting the headline finding is that the GPT-4o-judged recoverability rate (42.0%) matches the best Llama-based system's alias accuracy (41.4%) within 0.6 pp; the two estimates are produced by architecturally independent pipelines.

*Other limitations.* The benchmark covers English-language oral histories focused primarily on American experiences. The LLM-generated entity elision, while validated for naturalness and audited for possible leakage, is not a substitute for human-written implicit references. The deterministic alias-string filter screened 25,161 candidate EENs and removed 25 cases (0.10%), whereas the GPT-4o validation audit flagged 34 of 500 sampled cases (6.8%) as possible leakage under a broader judge-based rubric. Alias-aware evaluation still penalizes semantically correct predictions that use unregistered surface forms. Temperature controls were not run for Llama 3.1 8B CoT. QLoRA training used a maximum sequence length of 192 tokens, which truncates approximately 6% of test prompts.

## 8. Conclusion

We have extended implicit entity recognition, previously studied in short social-media text [9,10], to the domain of long-form reminiscence narratives, formalizing the non-locality property that distinguishes this setting and empirically validating it through a sentence-level ablation showing a 20.6pp accuracy gap between full-text and single-sentence inference. We release IRC-Bench, a benchmark of 25,136 samples spanning 12,337 Wikidata-linked entities from 1,994 oral history transcripts across 11 thematic domains. Our systematic evaluation across 19 experimental configurations reveals eight key findings.

First, fine-tuning is the single most impactful intervention. QLoRA-adapted Llama 3.1 8B achieves 38.94% exact match (51.59% Jaccard), nearly tripling the base model's zero-shot performance and surpassing GPT-4o few-shot by 7.32 percentage points, despite the entity-level split ensuring zero overlap with training entities. In the closed-world setting, DPR fine-tuning of BGE-base more than doubles Hit@1 from 16.64% to 35.38%, with the gold entity appearing in the top-10 for 71.49% of queries.

Second, chain-of-thought prompting degrades smaller models (by 4.51 to 7.70 pp), while temperature control experiments reveal that for GPT-4o, the observed CoT penalty is largely attributable to the higher sampling temperature rather than the reasoning structure itself. In all cases, CoT fails to exceed zero-shot performance, confirming that implicit entity recognition requires holistic pattern matching rather than sequential reasoning. Third, retrieval-augmented generation underperforms direct LLM inference due to the non-locality of implicit cues. Fourth, model scale is the dominant factor in zero-shot open-world accuracy, with performance spanning from 13.92% (Llama 3.1 8B) to 27.02% (GPT-4o) in exact match. Fifth, entity descriptions are consistently the best representation for dense retrieval, outperforming both entity names and Wikipedia lead sentences.

Future work should explore several promising directions: multi-modal implicit entity recognition incorporating audio features from the original recordings, cross-lingual benchmarks constructed from oral history archives in other languages, active learning approaches that combine fine-tuned DPR shortlists with fine-tuned LLM reranking (leveraging C5's 71.49% Hit@10), and the development of specialized architectures that explicitly model the non-locality property of implicit entity cues through structured attention over distributed text spans.

**Author Contributions:** Conceptualization, Y.A. and A.A.; methodology, Y.A. and A.A.; software, E.M.; validation, Y.A., E.M., and A.A.; formal analysis, Y.A., E.M., and A.A.; data curation, E.M. and A.A.; writing—original draft preparation, Y.A. and A.A.; writing—review and editing, Y.A. and A.A.; visualization, Y.A., E.M., and A.A.; supervision, Y.A. and A.A.; project administration, Y.A. All authors have read and agreed to the published version of the manuscript

**Funding:** This research received no external funding

**Data Availability Statement:** The IRC-Bench dataset, source code, experimental scripts, prompts, evaluation tools, and result files supporting the findings of this study are publicly available in the project repository at https://github.com/ApartsinProjects/ImplicitEntities. The benchmark annotations are released under the license specified in the repository. Source oral-history transcripts are publicly available from their respective archives, with provenance information provided in the repository.

**Conflicts of Interest:** The authors declare no conflicts of interest.

## Abbreviations

The following abbreviations are used in this manuscript:

| | |
|---|---|
| BAAI | Beijing Academy of Artificial Intelligence |
| BGE | BAAI General Embedding |
| BiLSTM-CRF | Bidirectional long short-term memory with conditional random fields |
| BLINK | Bi-encoder Linker |
| C1-C6 | Closed-world experimental configurations 1-6 |
| CoT | Chain-of-thought |
| DPR | Dense passage retrieval |
| EEN | Entity-elided narrative |
| EGN | Entity-grounded narrative |
| EL | Entity linking |
| FS | Few-shot |
| GENRE | Generative Entity Retrieval |
| IRC-Bench | Implicit Reminiscence Context Benchmark |
| LoRA | Low-Rank Adaptation |
| MNRL | Multiple Negatives Ranking Loss |
| MRR | Mean Reciprocal Rank |
| MTEB | Massive Text Embedding Benchmark |
| NER | Named Entity Recognition |
| NF4 | NormalFloat 4-bit |
| O1-O13 | Open-world experimental configurations 1-13 |
| QID | Wikidata identifier |
| QLoRA | Quantized Low-Rank Adaptation |
| RAG | Retrieval-augmented generation |
| W2NER | Word-word relation classification for named entity recognition |
| ZS | Zero-shot |

## Appendix A: Prompt Templates

*Appendix A.1. Zero-Shot Prompt*

System message:

> You are an entity recognition expert. Given a text that implicitly references a named entity without mentioning it, identify what entity is being referenced.

User message:

> What named entity is implicitly referenced in this text? The entity is never mentioned by name.
>
> Text: "{text}"
>
> Think about the contextual cues (dates, places, events, people, roles) and identify the specific named entity being referenced.
>
> Answer with ONLY the entity name (canonical Wikipedia name), nothing else.

Parameters: temperature=0.0, max_tokens=100

*Appendix A.2. Few-Shot Prompt*

System message:

> You are an entity recognition expert. Given a text that implicitly references a named entity without mentioning it, identify what entity is being referenced.

User message:

> What named entity is implicitly referenced in this text? The entity is never mentioned by name.
>
> Examples: Text: "I remember that Sunday morning in December '41. We were listening to the radio when the news broke about the attack on the naval base in Hawaii. That's when everything changed." Entity: Attack on Pearl Harbor

Text: "I enlisted right out of high school and went to boot camp in San Diego. As an aircraft mechanic, I was sent to the Pacific." Entity: United States Marine Corps

Text: "After the surrender, we flew into the main islands. I landed in the bay and spent six months there for the occupation. The capital was flattened by the B-29s." Entity: Tokyo

Text: "In late 1941, I was set to ship out from San Francisco. A friend ran up saying they're bombing the base in Hawaii." Entity: Attack on Pearl Harbor Text: "Growing up in that bustling metropolis with towering skyscrapers, I was immersed in a vibrant culture." Entity: New York City

Now identify the entity in this text:

Text: "{text}"

Answer with ONLY the entity name (canonical Wikipedia name), nothing else.

Parameters: temperature=0.0, max_tokens=100

*Appendix A.3. Chain-of-Thought Prompt*

System message:

You are an entity recognition expert. Think step by step.

User message:

What named entity is implicitly referenced in this text? The entity is never mentioned by name.

Text: "{text}"

Think step by step:

1. What contextual cues are present? (dates, places, events, people, roles)

2. What type of entity do these cues suggest? (Person, Place, Organization, Event)

3. What specific named entity matches ALL these cues?

Reasoning: [your step-by-step analysis]

Entity: [canonical Wikipedia name]

Parameters: temperature=0.7, max_tokens=300

*Appendix A.4. RAG Prompt*

User message (no system message):

This text implicitly references a named entity without naming it. Based on the contextual cues, which candidate is most likely?

Text: "{text}"

Candidates:

1. {candidate_1} - {description_1}

2. {candidate_2} - {description_2}

3. {candidate_3} - {description_3}

4. {candidate_4} - {description_4}

5. {candidate_5} - {description_5}

If none match well, suggest a better entity.

Answer: [number]. [entity name]

Parameters: temperature=0.7, max_tokens=50

*Appendix A.5. QLoRA Fine-tuning Prompt (O10)*

<|begin_of_text|>

<|start_header_id|>system<|end_header_id|>

You identify implicitly referenced entities.<|eot_id|> <|start_header_id|>user<|end_header_id|>

What entity is implicitly referenced? Answer with only the entity name.

Text: {implicit_text}<|eot_id|> <|start_header_id|>assistant<|end_header_id|>

{entity}<|eot_id|>

Parameters: greedy decoding, max_new_tokens=30

*Appendix A.6. Pipeline Stage 2: NER + Wikidata-Title Prompt*

Used by Stage 2 of the corpus-construction pipeline to extract entities and their canonical Wikipedia titles. Model: GPT-4.1-mini, temperature 0.0, max_tokens 4000.

System message:

You are a named entity recognition expert. Output ONLY a valid JSON array. No markdown, no code fences, no explanation.

User message:

Extract ALL notable named entities from this oral history transcript. For each entity provide: - id: sequential number (1, 2, 3...) - entity: canonical name (as it appears on Wikipedia) - type: exactly one of: Person, Place, Organization, Event, Work, Military_Unit - surface_forms: array of exact text spans from the transcript that reference this entity (include all variations/mentions) - wikipedia_title: Wikipedia article title (best guess, use canonical English Wikipedia title) RULES: 1. Only include entities that would have a Wikipedia article 2. Include all notable: people, places, organizations, events, military units, cultural works 3. Do NOT include the interviewer or respondent themselves (unless they are a public figure being interviewed ABOUT their public role) 4. Do NOT include generic terms (army, soldier, war, city, country without specific name) 5. Include historical events with specific names (Pearl Harbor, D-Day, 9/11) 6. For places: include specific named locations (cities, bases, landmarks), not generic ("the beach", "my house") 7. Surface forms: list ALL different ways the entity is mentioned in the text Output ONLY a JSON array: [ {"id": 1, "entity": "Pearl Harbor", "type": "Event", "surface_forms": ["Pearl Harbor", "the attack on Pearl Harbor"], "wikipedia_title": "Attack_on_Pearl_Harbor"}, {"id": 2, "entity": "United States Marine Corps", "type": "Organization", "surface_forms": ["the Marines", "Marine Corps", "USMC"], "wikipedia_title": "United_States_Marine_Corps"} ] Transcript: {text}

*Appendix A.7. Pipeline Stage 3: Entity-Grounded Summary Prompt*

Used by Stage 3 to generate first-person summaries that mention each entity by name while preserving contextual cues. Model: GPT-4.1-mini, temperature 0.3, max_tokens 4000.

System message:

You are a research assistant creating entity-focused summaries from oral history transcripts. Output ONLY valid JSON. No markdown, no code fences.

User message:

Given this oral history transcript and a list of entities found in it, generate a short first-person summary for EACH entity.

Each summary should:

1. Be 3-5 sentences, written as if the respondent is telling the story

2. Mention the entity BY NAME naturally

3. Include contextual cues: dates, places, co-occurring events, roles, people

4. Capture the relationship between the respondent and the entity

5. Be factually grounded in the transcript (no invented details)

Entities to summarize:

{entities}

Transcript:

{text}

Output a JSON array with one object per entity:

```
[
  {
    "entity_id": 1,
    "entity": "Pearl Harbor",
    "summary": "I remember when Pearl Harbor was attacked. It was December 7th, 1941, and I was listening to the radio...",
    "cues": ["December 7th, 1941", "radio", "attack", "Hawaii"]
  }
]
```

Generate summaries for ALL entities listed above. Output ONLY the JSON array.

*Appendix A.8. Pipeline Stage 4: Entity Elision (Implicit Rewrite) Prompt*

Used by Stage 4 to rewrite each summary with the entity name removed while preserving every contextual cue. Model: GPT-4.1-mini, temperature 0.3, max_tokens 4000.

System message:

You rewrite text to remove entity names while preserving style and content. Output ONLY valid JSON. No markdown, no code fences.

User message:

Rewrite each summary so the target entity is NEVER mentioned by name, abbreviation, or obvious synonym.

RULES:

1. Remove ALL direct references to the entity name

2. Keep the first-person voice and speaking style

3. Keep all contextual cues: dates, places, co-occurring events, roles, people

4. Replace the entity name with natural contextual descriptions (not verbose)

5. Do NOT add or invent new information

6. Keep the same length (3-5 sentences)

7. The rewritten text should still allow a knowledgeable reader to identify the entity

Summaries to rewrite:

{summaries}

Output a JSON array with the rewritten versions:

[

```
    {
        "entity_id": 1,
        "entity": "Pearl Harbor",
        "implicit_text": "I remember when it happened. It was December 7th, 1941, and we were listening to a football game on the radio when the news broke about the attack on the naval base in Hawaii..."
    }
]
Rewrite ALL summaries. Output ONLY the JSON array.
```

## Appendix B: Training Hyperparameters

*Appendix B.1. DPR (Dense Passage Retrieval) Fine-tuning*

**Table B.1.** DPR (Dense Passage Retrieval) Fine-tuning

| Base model | BAAI/bge-base-en-v1.5 |
|---|---|
| Model parameters | ~110M |
| Embedding dimension | 768 |
| Training examples | 17,971 |
| Epochs | 3 |
| Batch size | 48 |
| Learning rate | 2e-5 |
| Warmup steps | 100 |
| Loss function | MultipleNegativesRankingLoss (MNRL) |
| Optimizer | AdamW |
| Mixed precision | FP16 (AMP) |
| Negatives | In-batch (47 negatives per sample) |
| Random seed | 42 |

Three separate models are trained, one for each entity representation (name, description, wiki). All use identical hyperparameters.

*Appendix B.2. QLoRA (O10) Fine-tuning*

**Table B.2.** QLoRA (O10) Fine-tuning

| Base model | meta-llama/Llama-3.1-8B-Instruct |
|---|---|
| Model parameters | ~8B (base); ~6.5M trainable (LoRA) |
| Quantization | 4-bit NormalFloat (NF4) |
| Compute dtype | bfloat16 |
| LoRA rank (r) | 16 |
| LoRA alpha ($\alpha$) | 32 |
| LoRA dropout | 0.05 |
| Target modules | q_proj, v_proj, k_proj, o_proj |
| Training examples | 17,971 |
| Epochs | 2 |
| Per-device batch size | 48 |
| Gradient accumulation | 1 (effective batch: 48) |

| Learning rate | 2e-4 |
|---|---|
| Max sequence length | 192 tokens |
| Warmup steps | 50 |
| Precision | bfloat16 |
| Validation samples | 500 (subset of dev) |
| Framework | TRL SFTTrainer + PEFT |

*Appendix B.3. Quality-Evaluation Judge Prompt*

Used by §3.8 to validate Stage-4 outputs on 500 test-partition EENs. The judge (GPT-4o, temperature 0.0, max_tokens 200) receives both the entity-grounded narrative (EGN) and the entity-elided narrative (EEN) and rates four dimensions. The verbatim user-message template is reproduced below; all 500 raw judgments are released with the benchmark.

You are evaluating the quality of an implicit entity recognition sample. A human narrator originally described an event/person/place by name (the Entity-Grounded Narrative). The entity name was then removed to create the Entity-Elided Narrative (EEN), preserving only contextual cues.

Your task: evaluate the EEN quality on four dimensions.

**Entity:** {entity_name} ({entity_type})

**EEN (entity name removed):** "{een_text}"

**EGN (original with entity name):** "{egn_text}"

Rate each dimension:

1. **Naturalness** (1-5): Does the EEN read like something a person would naturally say? (1=very awkward/robotic, 5=completely natural first-person speech)

2. **Leakage** (yes/no): Does the entity name (or an obvious alias) still appear verbatim in the EEN?

3. **Cue sufficiency** (1-5): Are there enough contextual cues in the EEN for a knowledgeable human to identify the entity? (1=impossible, 3=possible with expertise, 5=obvious)

4. **Recoverability** (yes/probably/unlikely/no): Could you identify the entity from the EEN alone, without seeing the EGN?

Respond in exactly this format:

Naturalness: [1-5]

Leakage: [yes/no]

Cue sufficiency: [1-5]

Recoverability: [yes/probably/unlikely/no]

Brief explanation: [one sentence]

## Appendix C: Example Predictions

We present five correct and five incorrect predictions from GPT-4o few-shot (O2) and selected O10 predictions, to illustrate the task characteristics and failure modes.

*Appendix C.1. Correct Predictions (O2: GPT-4o Few-shot)*

**Correct Example 1**

*"I studied at a major public university in Northern California, where I was part of the Design Department. During my time there, I combined academic courses with art classes, focusing on three-dimensional design..."*

**Gold:** University of California, Berkeley | **Prediction:** University of California, Berkeley ✓

**Correct Example 2**

*"I was born in the capital city of Germany in the early 1930s. It was a turbulent time as the political climate was rapidly changing. My family decided to leave that city in 1938 to escape the dangers posed by the Nazi regime. That move shaped much of my early life and future."*

**Gold:** Berlin | **Prediction:** Berlin ✓

**Correct Example 3**

*"When I first came to America, I worked in a Pacific island territory for eight months on a sugar plantation. I was only 15 years old and worked under a Chinese boss for $18 a month..."*

**Gold:** Hawaii | **Prediction:** Hawaii ✓

**Correct Example 4**

*"My grandfather was a teenager during the major 1950s political upheaval in our Caribbean homeland and once found a journal from someone fighting with the revolutionary leader. That period was filled with fear for my family and community. The uprising brought about communism, which had some positive effects like high literacy rates, but also caused extreme poverty and suffering. The memories of that era shape how older immigrants from that island view politics in the United States today."*

**Gold:** Cuban Revolution | **Prediction:** Cuban Revolution ✓

**Correct Example 5**

*"He and I were close when he was Senate majority leader, and he was very cordial to me when I first came to the Senate. He gave me important committee assignments, including chairing the Calendar Committee and seats on the Agricultural and Finance Committees. He was probably the most able majority leader in history, knowing the Senate's personalities and how to motivate them. As President, he overcommitted on social programs, which I believe contributed to the huge deficits we face today."*

**Gold:** Lyndon B. Johnson | **Prediction:** Lyndon B. Johnson ✓

These correct examples demonstrate cases where geographic cues ("capital city of Germany," "Pacific island territory"), temporal markers ("1950s," "Senate majority leader"), and contextual details (the Nazi regime, sugar plantations, Caribbean communism) are sufficiently distinctive for the model to identify entities across Place, Event, and Person types.

*Appendix C.2. Incorrect Predictions (O2: GPT-4o Few-shot)*

**Incorrect Example 1: Wrong type**

*"My great-grandfather left Lithuania in the early 1900s to escape oppression and seek a better life in America. He arrived before World War I and worked hard to establish himself, eventually sending for his family..."*

**Gold**: Solomon Goodman | **Prediction**: Lithuanian Jews ✗

The model predicted a group/category rather than the specific individual being described.

**Incorrect Example 2: Wrong type**

*"I once kept a newspaper clipping of a write-up about a historic estate by a famous 19th-century author, but unfortunately, I have misplaced it. The author's writing gave me some insight into the estate..."*

**Gold**: Harriet Beecher Stowe | **Prediction:** Monticello ✗

The model focused on the "historic estate" cue rather than the "famous 19th-century author" cue.

**Incorrect Example 3: Wrong type**

*"The priest who taught me algebra and later became the bishop of a diocese in eastern Washington questioned my presence in his advanced algebra class because I lacked the necessary background..."*

**Gold**: Bishop of Spokane | **Prediction**: West Point ✗

The model produced a completely unrelated entity, likely confusing the religious context.

**Incorrect Example 4: Same-type, related**

*"Remote healthcare delivery became a critical part of how we provided care during the viral outbreak. Initially, we relied on phone calls, but within weeks, our organization quickly implemented video..."*

**Gold**: Telehealth | **Prediction**: COVID-19 pandemic ✗

The model identified the correct general domain but predicted the contextual event rather than the practice being described.

**Incorrect Example 5: Same-type, near miss**

*"He was my Ph.D. advisor at the California university starting in 1956. He was a brilliant economist who later won the Nobel Prize, and studying under him greatly influenced my academic development. His mentorship helped shape my approach to economics and game theory."*

**Gold**: Kenneth Arrow | **Prediction**: John Forbes Nash Jr. ✗

The model predicted a Nobel laureate economist associated with game theory, but confused the advisor (Arrow, at Stanford) with another famous figure in the same field.

These errors illustrate the principal challenges of implicit entity recognition: distinguishing the referenced entity from related contextual entities (Examples 4, 5), resolving references to obscure individuals (Examples 1, 3), and focusing on the correct cue among multiple competing signals (Example 2). Example 5 is particularly instructive: both Kenneth Arrow and John Nash are Nobel laureate economists linked to game theory, but the cues (Ph.D. advisor, California, 1956) point specifically to Arrow at Stanford.